%% file: neurips_2025.tex
\documentclass{article}

\PassOptionsToPackage{numbers, compress}{natbib}

\usepackage[preprint]{neurips_2025}

\usepackage[utf8]{inputenc} 
\usepackage[T1]{fontenc}    
\usepackage{hyperref}       
\usepackage{url}            
\usepackage{booktabs}       
\usepackage{amsfonts}       
\usepackage{nicefrac}       
\usepackage{microtype}      
\usepackage{xcolor}         

\usepackage{algorithmic} 
\usepackage{algorithm} 
\usepackage{amsmath}
\usepackage{multirow}
\usepackage{subfigure}
\usepackage{caption}

\usepackage{tcolorbox}
\usepackage{tikz}

\usepackage{xcolor}
\usepackage{colortbl}
\usepackage{makecell}

\usepackage{listings}

\newcommand{\customcode}[2][]{ 
    \begin{minipage}{1\textwidth}
    \lstset{
        backgroundcolor=\color{white}, 
        commentstyle=\color{gray}, 
        keywordstyle=\color{blue}, 
        stringstyle=\color{orange}, 
        basicstyle=\ttfamily\footnotesize\bfseries, 
        breakatwhitespace=false,
        breaklines=true,
        keepspaces=true,
        numbers=left,
        numbersep=8pt, 
        numberstyle=\scriptsize\color{gray}, 
        showspaces=false,
        showstringspaces=false,
        showtabs=false,
        tabsize=2,
        xleftmargin=12pt, 
        frame=single, 
        rulecolor=\color{black}, 
    }
    \lstinputlisting[firstnumber=#1,language=Python]{#2} 
    \end{minipage}
}

\newtcolorbox{dialogbox}[1][]{
  arc=4mm,
  colback=lightgray!15,
  colframe=cyan!40!black,
  rounded corners,
  boxrule=1.5pt,
  fonttitle=\sffamily\bfseries,
  coltitle=white,
  toptitle=2mm,
  bottomtitle=2mm,
  title=#1, 
  width=1\linewidth,
  center,
}

\newtcolorbox{dialogbox1}[1][]{
  arc=4mm,
  colback=lightgray!15,
  colframe=teal!70!black,
  rounded corners,
  boxrule=1.5pt,
  fonttitle=\sffamily\bfseries,
  coltitle=white,
  toptitle=2mm,
  bottomtitle=2mm,
  title=#1, 
  width=1\linewidth,
  center,
}

\usepackage{listings}
\title{ARS: Automatic Routing Solver with Large Language Models}

\makeatletter
\renewcommand{\thanks}[1]{%
  \footnotemark[1] 
  \protected@xdef\@thanks{\@thanks\protect\footnotetext[1]{#1}}%
}
\makeatother

\author{
  Kai Li$^1$
  \quad Fei Liu$^{2}$
  \quad Zhenkun Wang$^{1}$\thanks{Corresponding author}
  \quad Xialiang Tong$^{3}$
  \quad Xiongwei Han$^{3}$
  \\ \textbf{Mingxuan Yuan$^{3}$
  \quad Qingfu Zhang$^{2}$}\\
    $^1$ Southern University of Science and Technology\\
    $^2$ City University of Hong Kong \\
    $^3$ Huawei Noah’s Ark Lab \\
  \texttt{12332664@mail.sustech.edu.cn, fliu36-c@my.cityu.edu.hk, wangzhenkun90@gmail.com} \\
  \texttt{\{tongxialiang, hanxiongwei, Yuan.Mingxuan\}@huawei.com} \\
  \texttt{qingfu.zhang@cityu.edu.hk}
  }

\begin{document}

\maketitle

\input{0-Abstract}
\input{1-Introduction}
\input{2-Background}

\input{3-Method}

\input{4-Benchmark}

\input{5-Experiments}

\input{6-Conclusion}

{
\small

\bibliographystyle{unsrtnat}
\bibliography{neurips_2025}        



}

\newpage

\appendix


\input{10-Appendix}

\end{document}

%% file: 0-Abstract.tex
\begin{abstract}

Real-world Vehicle Routing Problems (VRPs) are characterized by a variety of practical constraints, making manual solver design both knowledge-intensive and time-consuming. Although there is increasing effort in automating the design of routing solvers, existing research has explored only a limited array of VRP variants and fails to adequately address the complex and prevalent constraints encountered in real-world situations. We propose Automatic Routing Solver (ARS), which leverages Large Language Model (LLM) agents to enhance a backbone metaheuristic framework. ARS automatically generates constraint-aware heuristic code using problem descriptions and representative constraints selected from a database, offering greater flexibility and interpretability than existing methods without relying on cumbersome modeling rules. Alongside ARS, we introduce RoutBench~\footnote{https://github.com/Ahalikai/ARS-Routbench/}, a benchmark comprising 1,000 VRP variants derived from 24 attributes, designed to rigorously evaluate the effectiveness of automatic routing solvers in handling VRPs with diverse practical constraints. We evaluate ARS against seven LLM-based methods on both common VRPs and RoutBench. Experimental results show that ARS achieves a success rate of over 90\% on common VRPs and over 60\% on RoutBench, outperforming other methods by at least 30\% in success rate. In terms of solving performance, ARS also significantly surpasses general-purpose solvers, demonstrating higher efficiency across many VRP variants.

\end{abstract}

%% file: 1-Introduction.tex
\section{Introduction}\label{Introduction}

The Vehicle Routing Problem (VRP) is a fundamental Combinatorial Optimization Problem (COP) that plays a critical role in logistics, transportation, manufacturing, retail distribution, and delivery planning~\citep{toth2014vehicle, liu2023heuristics}.
In these scenarios, the objective of VRPs is to efficiently allocate and plan vehicle routes to meet various requirements while minimizing overall routing costs. 
These requirements often include constraints such as vehicle capacities, time windows, and duration limits, resulting in numerous variants of VRPs in practical applications~\citep{braekers2016vehicle}.
However, existing heuristics are usually problem-specific. When the problem changes slightly (e.g., a minor modification to the requirements), a lot of effort is required for experts to redesign the heuristic to make it effective in solving the new problems~\citep{vidal2013hybrid,rabbouch2021efficient,errami2023vrpsolvereasy}.

Large Language Models (LLMs) have shown powerful reasoning and code-generation capabilities~\citep{chen2021evaluating, austin2021program, li2023starcoder}. By integrating these functionalities, users can express their specific requirements in natural language, enabling the models to automatically design algorithms to address VRPs~\citep{liu2024systematic}. Most existing works primarily leverage the LLM to solve a small number of standard VRPs~\citep{jiang2024unco, huang2024words} and cannot be applied to complex VRPs.
Some recent studies have employed LLMs to formulate routing problems as integer programming models and solve them using general-purpose solvers~\citep{xiao2023chain, zhang2024solving}. However, these approaches may face challenges in adhering to standard modeling practices, and the solvers often exhibit low efficiency, limiting their ability to handle complex real-world constraints.

To tackle these challenges, this paper proposes a framework that uses a heuristic algorithm, developed with the assistance of LLMs, to solve the VRP variants with complex constraints.
Our contributions are summarized as follows:

\begin{itemize}

    \item We propose ARS, a framework designed to automatically create constraint-aware heuristics based on the problem description, which enhances a backbone heuristic algorithm for route optimization, offering an adaptive framework to address the diverse routing problems expressed in natural language.

    \item We introduce RoutBench, a benchmark with 1,000 VRP variants derived from 24 VRP constraints.
    Each variant in RoutBench is equipped with a detailed problem description, instance data, and validation code, enabling the evaluation of various routing solvers' effectiveness in handling diverse VRP constraints.
    \item We comprehensively validate our approach on commonly-used problems and RoutBench. The results show that ARS can automatically handle 91.67\% of common problems, achieving a significant improvement across all benchmarks compared to other LLM-based methods and general solvers.

\end{itemize}

%% file: 2-Background.tex
\section{Problem Formulation}

Vehicle Routing Problems (VRPs) involve optimizing the routes and schedules of a fleet of vehicles delivering goods or services to various locations, aiming to minimize costs while satisfying constraints like delivery windows and vehicle capacities.
The VRP variants can be mathematically described as optimization problems on a graph $\mathcal{G} = (\mathcal{V}, \mathcal{E})$ where nodes $\mathcal{V} = \{0, 1, \ldots, n\}$ represent depot $0$ and locations $\{1,\ldots, n\}$, and edges represent the possible routes between these nodes $\mathcal{E}=\{e_{ij}, i,j\in \mathcal{V} \}$, each of them is assigned with a cost $c_{ij}$. 
The mathematical representation is given by:

\begin{equation} \label{eqn:VRP}
    \begin{aligned}
        \min \sum_{i \in V} \sum_{j \in V} c_{ij} x_{ij}, \\
        \text{subject to} \quad \mathbf{x} \in C,
    \end{aligned}
\end{equation}
where $\mathbf{x} = \{x_{ij} \mid i, j \in \mathcal{V},\ i \neq j\}$
represents the set of decision variables, \(x_{ij}\) is a binary variable that indicates whether the route from \(i\) to \(j\) is used.
The feasible solution space \(C\) is defined by constraints. In this paper, we consider VRPs with a variety of real-world constraints such as vehicle capacity, travel distance, and time windows, thereby extending the basic VRP, as seen in the Capacitated VRP (CVRP)~\citep{toth2014vehicle} and the VRP with Time Windows (VRPTW)~\citep{solomon1987algorithms}.
Moreover, new constraints often emerge in real-world scenarios. For example, VRP variants related to vehicle capacity include Heterogeneous VRP (HVRP), which considers vehicles with different capacities~\citep{lai2016tabu}, Multi-Product VRP (MPVRP), which addresses the need to transport multiple types of products~\citep{yuceer1997multi}, and dynamic demands~\citep{powell1986stochastic}. VRP variants that incorporate these real-world constraints are more prevalent and practically significant in real-world applications.

However, current methods focus on a limited range of problems and do not sufficiently address the complex and diverse constraints present in real-world scenarios. To bridge this gap and further the development of automated solutions for practical VRP variants, this paper introduces a benchmark for VRPs featuring various complex yet practical constraints. Additionally, we propose a general automatic routing solver enhanced by a large language model to effectively manage these constraints.

%% file: 3-Method.tex
\section{Automatic Routing Solver}\label{Method}

\begin{figure*}[t]
    \centering
    \includegraphics[width=1\linewidth]{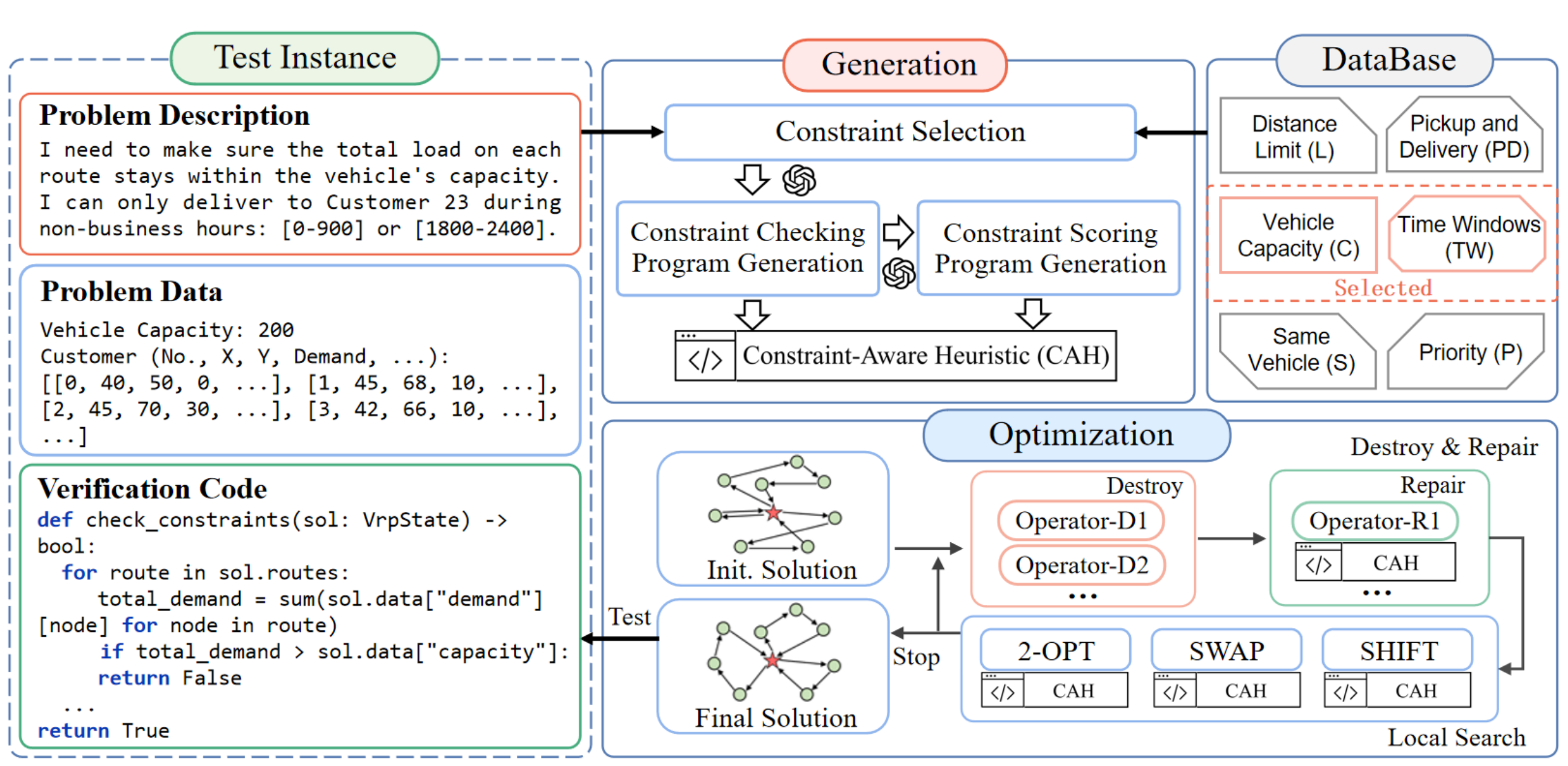}
    \caption{Overview of the proposed ARS framework. The left side of the figure shows the test instance, including the problem description, corresponding data, and validation code for result verification. The right side comprises the database, generation module, and VRP solver. The generation module selects relevant constraints from the database and generates constraint-aware heuristics for the VRP solver to address different VRP variants.
    }
    \label{fig:vrps}
\end{figure*}

Given the problem description in natural language format and the instance data for any VRP variants with one or more constraints, our proposed Automatic Routing Solver (ARS) can automatically generate the constraint-aware heuristic to augment the backbone heuristic solver.
ARS consists of three key components: 1) Pre-defined Database, 2) Constraint-aware heuristic generation, and 3) Augmented heuristic solver.

\subsection{Database}
We build a database, denoted as $D$, with several representative fundamental constraints to provide additional guidance for LLM-driven constraint-aware heuristic generation. Specifically, database $D$ includes a basic VRP information $(I_0, C_0)$ (without additional constraints) and six representative constraints $(I_k, C_k), k = 1, \dots, 6$, each corresponds to a distinct representative constraint: Vehicle Capacity, Distance Limit, Time Windows, Pickup and Delivery, Same Vehicle, and Priority.

Each constraint example $(I_k, C_k)$ consists of two parts:

\begin{itemize}

    \item $I_k$: The problem description. The natural language description of the constraint.
    
    \item $C_k$: The constraint feasibility checking program. It checks whether a solution belongs to the feasible solution space described by $I_k$. The program is given a solution and returns "True" if the corresponding constraint is satisfied. 
\end{itemize}

\subsection{Constraint-Aware Heuristic}

We first select relative constraints from the database. Then we sequentially generate the constraint checking and scoring programs for the target VRP variants, given the selected constraints and the problem description. Finally, the constraint-aware heuristic is generated based on the designed constraint checking and scoring programs.

\subsubsection{Constraint Selection}

Given an input problem description $I$, we instruct LLM agents to automatically select a subset of relevant constraints $S$ from the database $D$ to provide reference information for subsequent constraint program generation. There can be two cases. In the first case, LLM agents select one or more relative constraint examples. In the second case, if no constraints are recognized as related to the input $I$, the base case $(I_0, C_0)$, with no additional constraints, is selected.

This step is similar to Retrieval-Augmented Generation (RAG) methods that retrieve information from external sources~\citep{zhang2023repocoder, jiang2025droc}, while our candidate reference information is a pre-defined database with representative constraints.

\subsubsection{Constraint Checking Program Generation}

This approach works by taking a set of relevant constraints, denoted as $S$. For each constraint $C_k$ within $S$, we incorporate specific modifications, $\Delta C_k(I)$, based on the problem description $I$. This process results in the creation of new, customized constraints $C_{new}$, which are better suited to address the specific requirements of the problem. This method offers two main benefits:

\begin{itemize}
    \item User-Centric Design: It aligns with how users typically work by enhancing existing constraints. This allows users to refine their specific requirements without the need to develop a complete problem definition from the beginning.
    \item Efficient Processing by LLMs: It helps LLM agents focus on these new, tailored constraints. This reduces unnecessary complexity and enhances the relevance and accuracy of the solutions provided by the models.
\end{itemize}

\subsubsection{Constraint Scoring Program Generation}

In practice, heuristic algorithms operate within a variable space that is typically much larger than the feasible solution space. This discrepancy presents significant challenges in identifying feasible solutions. To address this, heuristic methods may permit the presence of infeasible solutions during the search process, as this allows for the evaluation of potential improvements in solution quality \citep{deb2000efficient, maximo2021hybrid}.

Therefore, to effectively integrate $C_{\text{new}}$ into the solution process, we utilize LLM agents to generate a violation score function guided by the constraint checking program. This score function quantifies the degree of constraint violation, thereby establishing a method for handling constraints and identifying high-quality solutions. It aids in systematically assessing and managing constraint violations, facilitating the search for feasible and high-quality solutions within the expansive variable space.

\subsubsection{Constraint-Aware Heuristic Generation}

We present the Constraint-Aware Heuristic (CAH) based on the constraint checking (Checker) and scoring programs (Scorer). As seen in Algorithm~\ref{Alg:constraint_handling}, the constraint handling method evaluates whether a new solution ${s}_{new}$ improves upon an old solution ${s}_{old}$. 
It first verifies whether ${s}_{old}$ is feasible (line 1). If ${s}_{old}$ is feasible, it then verifies whether ${s}_{new}$ is feasible and has a smaller travel distance (line 2). If both conditions are satisfied, ${s}_{new}$ is accepted.  
If ${s}_{old}$ is infeasible, but ${s}_{new}$ is either feasible or has a lower violation score than ${s}_{old}$ (line 5), ${s}_{new}$ is also accepted.
This approach allows infeasible solutions to evolve gradually toward feasibility while minimizing the overall travel distance.

\begin{algorithm}[]
\caption{Constraint-Aware Heuristic (CAH)}
\label{Alg:constraint_handling}
\centering
\begin{algorithmic}[1]
\REQUIRE New solution ${s}_{new}$, Old solution ${s}_{old}$

\IF{Checker(${s}_{old}$) is feasible}
    \IF{Checker(${s}_{new}$) is feasible \textbf{and} Cost(${s}_{new}$) $<$ Cost(${s}_{old}$)}
        \STATE \textbf{return true};
    \ENDIF
\ELSIF{Checker(${s}_{new}$) is feasible \textbf{or} Scorer(${s}_{new}$) $<$ Scorer(${s}_{old}$)}
        \STATE \textbf{return true};
\ENDIF
\STATE \textbf{return false};
\end{algorithmic}
\end{algorithm}

\subsection{Augmented Heuristic Solver}
The solver has a general single-point-based search backbone heuristic framework, which utilizes automatically generated constraint-aware heuristics to solve various VRP variants.
This backbone heuristic solver mainly consists of 1) destroy\&repair and 2) local search.

In the destroy phase, we employ multiple operators, including random removal and string removal~\citep{christiaens2020slack}, to selectively remove customers or partial routes from the current solution. The choice of destroy operators is determined using a roulette wheel selection mechanism, which assigns higher probabilities to operators that performed well in previous iterations. The repair phase reinserts removed customers into the solution using a greedy repair operator, aiming to construct a feasible solution with shorter routes.
Specific details of these operators are provided in Appendix~\ref{Appendix:Oper}.

Following the destroy and repair phases, the solution undergoes a local search process to further refine its quality. We utilize a set of local search operators, including 2-OPT~\citep{lin1965computer}, SWAP~\citep{osman1993metastrategy}, and SHIFT~\citep{rosenkrantz1977analysis}, and the best solution found among these operators is selected. To avoid premature convergence, the Record-to-Record Travel (RRT) criterion is applied~\citep{dueck1993new, santini2018comparison}, allowing the acceptance of slightly worse solutions within a predefined threshold, thus maintaining a balance between intensification and diversification. The entire process is iteratively repeated until a termination condition is met, such as a time limit.

Throughout the search, the constraint-aware heuristic plays a critical role in evaluating and guiding solutions. It is used in the recreate step and all local search operators. By effectively navigating infeasible regions and gradually improving solution feasibility, it ensures that the solver not only satisfies user requirements but also effectively minimizes the total route length.

%% file: 4-Benchmark.tex
\section{RoutBench}

Recent advancements in LLMs have opened up new possibilities for automatically generating routing solvers to address different VRP variants~\citep{xiao2023chain, chen2023teaching}.
However, they are typically evaluated on only a few dozen simple problems, leaving a significant gap in assessing their generalization ability. There is yet to be a VRP benchmark that can evaluate the generalization ability of these LLM-based methods, especially under complex yet common constraints in real-world scenarios.

\input{Tabs/VRP_Variants}

Thus, we propose RoutBench, a benchmark dataset that includes 1,000 VRP variants derived from 24 constraints. 
These constraints are chosen for their practical significance and theoretical challenges, as highlighted in Table~\ref{tab:VRP_Variants}. 
This design not only expands the test scale by two orders of magnitude but also provides an opportunity to evaluate algorithms on unseen VRPs.
If an algorithm can effectively solve these unseen problems, it demonstrates the potential to address new real-world challenges, better meeting practical application needs.

\begin{figure*}[t]
    \centering
    \begin{minipage}[]{0.49\textwidth}
        \centering
        \includegraphics[width=0.98\linewidth]{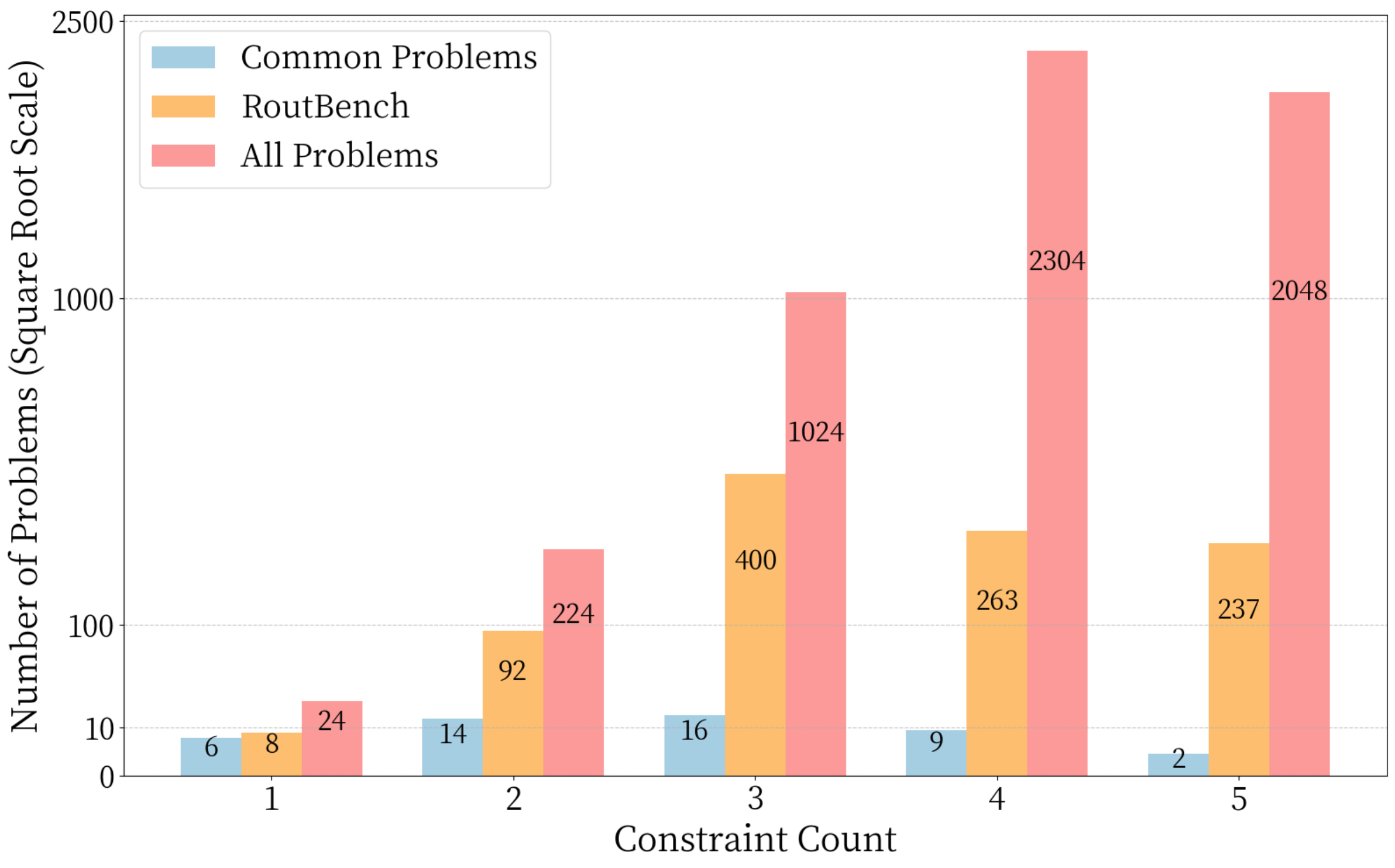}
        \caption{Analysis of problem distribution across common problems, RoutBench, and all problems.}
        \label{fig:Types}
    \end{minipage}
    \hfill 
    \begin{minipage}[]{0.49\textwidth}
        \centering
        \includegraphics[width=\linewidth]{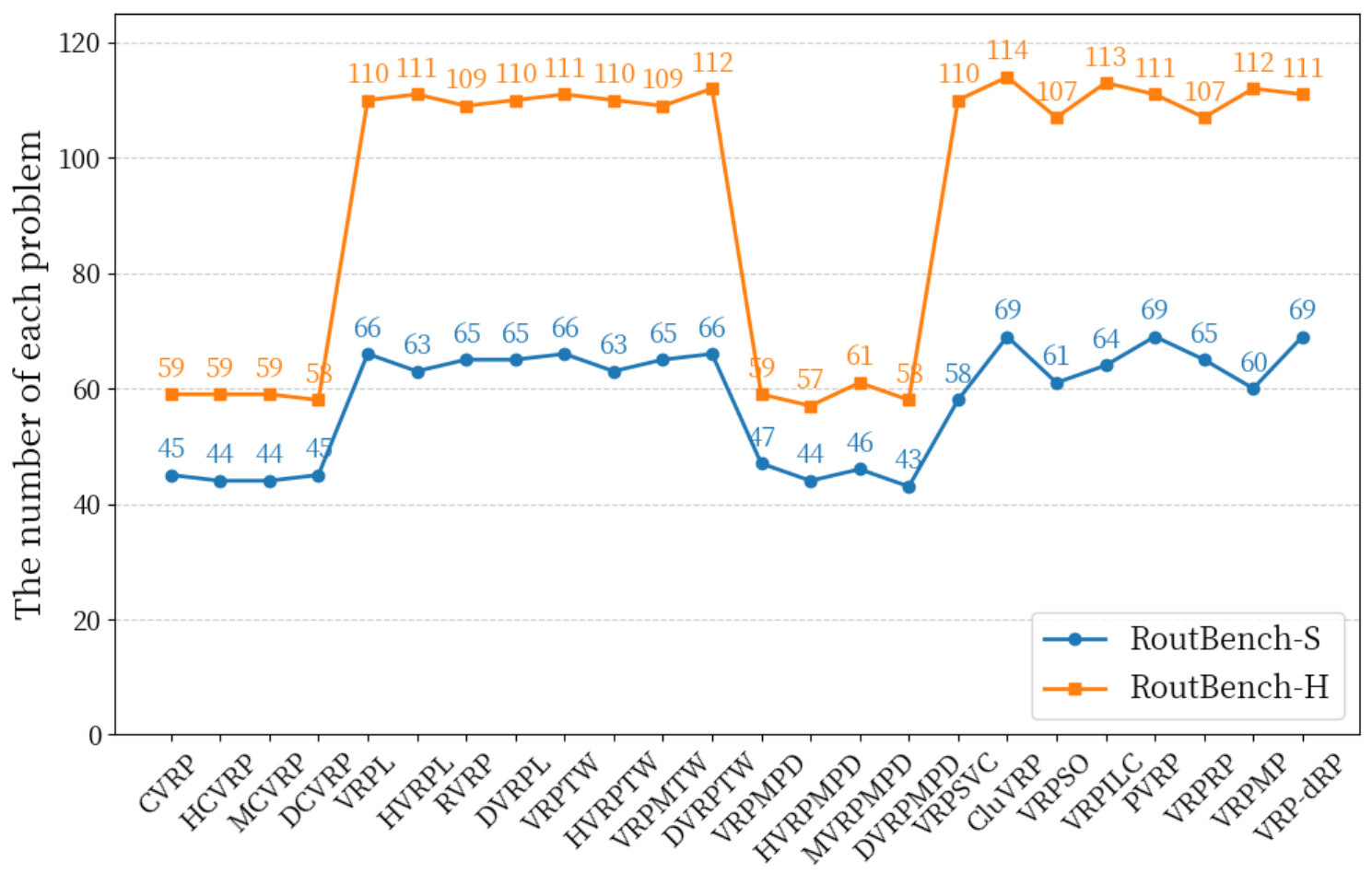}
        \caption{The frequency of constraint usage is analyzed for RoutBench-S and RoutBench-H.}
        \label{fig:problem}
    \end{minipage}
\end{figure*}

\subsection{Dataset Construction}

The RoutBench is constructed by combinations of six basic constraint types: Vehicle Capacity (C), Distance Limits (L), Time Windows (TW), Pickup and Delivery (PD), Same Vehicle (S), and Priority (P). We design four representative variants derived from each basic constraint type that incorporate variations such as heterogeneous vehicle fleets, multidimensional resource limits, dynamic changes, and others. To produce one problem instance, we first pick one basic constraint combination and then pick one real constraint variant for each chosen basic constraint type. Notice that Vehicle Capacity (C) and Pickup and Delivery (PD) are mutually exclusive, since PD is essentially a derivative of C. The total number of problem combinations is :

\begin{equation}
N_{\text{combinations}} = \sum_{k=1}^6 \binom{6}{k} \cdot 4^k,
\end{equation}
where \(\binom{6}{k}\) represents the number of ways to select \(k\) constraints from the six types, and \(4^k\) accounts for the four variations per constraint. After excluding combinations where vehicle capacity and pickup and delivery coexist, the total number of feasible combinations is reduced to 5624. From these, 1000 unique problem instances are uniformly sampled based on the order of all combinations, ensuring even coverage across the solution space. 
To balance complexity, RoutBench-S includes 500 problems with three or fewer constraints, while RoutBench-H consists of the remaining 500 problems with more than three constraints.

Each problem instance is comprised of three components: 1) the problem description, which is a natural language explanation of the problem; 2) the instance data, including the geometric positions of nodes and the constraint parameters, with data generated using the Solomon C103 dataset~\citep{solomon1987algorithms} as a base; and 3) the validation code, used to confirm whether a solution adheres to user requirements and satisfies all constraints. Problem sizes include 25, 50, and 100 nodes. All instances in RoutBench come with Best Known Solutions (BKS), with details provided in Appendix~\ref{10-A-BKS}.

The detailed descriptions of the 24 problem instances are provided in Table~\ref{table:24_prompt}. These examples illustrate the diversity of problem settings and serve as a representative subset of the dataset’s broader scope. By systematically combining constraints, leveraging validation mechanisms, and ensuring feasibility, RoutBench offers a diverse and reliable dataset for benchmarking VRP solvers.

\subsection{Analysis}

This section analyzes the distribution of problem types and complexities in the RoutBench dataset, focusing on the 48-problem subset and RoutBench, and their relationship to the full set of 5624 feasible problems.

The distribution of problems by the number of constraints is shown in Figure~\ref{fig:Types}.
The 48-problem subset consists of common problems, one is a simple VRP without any constraints, while the remaining 47 include one to five basic constraints.
In RoutBench, the distribution reflects the proportions of the full set of 5624 problems. Specifically, problems with three constraints are the most common in the dataset, as three-constraint problems dominate the total number of problems with three or fewer constraints. For more complex problems, those with four and five constraints appear in similar proportions, ensuring a balanced representation of high-complexity scenarios.

Figure~\ref{fig:problem} shows the distribution of problem types across RoutBench. Most problem types, such as VRPTW, HVRPL, and CluVRP, are well-represented. However, problems involving vehicle capacity constraints (e.g., CVRP, HCVRP) and pickup and delivery operations (e.g., VRPMPD, HVRPMPD) are less frequent. This is because these two categories are mutually exclusive, and the two types do not coexist in the dataset.

Overall, the RoutBench dataset achieves a diverse and balanced representation of problem types and complexities. The 48-problem subset provides a concise overview of simpler cases, while RoutBench captures a wide range of scenarios.

%% file: Tabs/VRP_Variants.tex
\begin{table}[ht]
\centering
\caption{A classification of common VRP variants is presented based on six constraint types, with abbreviations provided for both the constraints and VRP variants in parentheses. Each constraint type includes four distinct variants, with examples provided for each, resulting in a total of 24 constraints.}
\renewcommand{\arraystretch}{1.3}
\resizebox{1\textwidth}{!}{%
\begin{tabular}{clcl}
\toprule
Basic Constraints & \multicolumn{1}{c}{VRP Variants} & Basic Constraints & \multicolumn{1}{c}{VRP Variants} \\
\midrule
\multirow{4}{*}{\makecell{Vehicle \\ Capacity (C)}} & Capacitated VRP (CVRP)~\citep{vidal2022hybrid, luo2023neural} & \multirow{4}{*}{\makecell{Distance \\ Limit (L)}} & VRP with Distance Limit (VRPL)~\citep{laporte1985optimal, li1992distance} \\
& Heterogeneous CVRP (HCVRP)~\citep{lai2016tabu} & & Heterogeneous VRPL (HVRPL)~\citep{lee2021latent} \\
& Multi-Product VRP (MVRP)~\citep{yuceer1997multi} & & Recharging VRP (RVRP)~\citep{conrad2011recharging, erdougan2012green} \\
& Dynamic CVRP (DCVRP)~\citep{powell1986stochastic} & & Dynamic VRPL (DVRPL)~\citep{suzuki2011new, khouadjia2012comparative, qian2016fuel} \\
\midrule
\multirow{4}{*}{\makecell{Time \\ Windows (TW)}} & VRP with Time Windows (VRPTW)~\citep{solomon1987algorithms} & \multirow{4}{*}{\makecell{Pickup and \\ Delivery (PD)}} & VRP with Mixed Pickup and Delivery (VRPMPD)~\citep{avci2015adaptive} \\
& Heterogeneous VRPTW (HVRPTW)~\citep{ren2010multi, vidal2014unified} & & Heterogeneous VRPMPD (HVRPMPD)~\citep{avci2016hybrid} \\
& VRP with Multiple Time Windows (VRPMTW)~\citep{belhaiza2014hybrid} & & Multi-Product VRPMPD (MVRPMPD)~\citep{zhang2019multi} \\
& Dynamic VRPTW (DVRPTW)~\citep{ghiani2003real, pillac2013review} & & Dynamic VRPMPD (DVRPMPD)~\citep{gendreau2006neighborhood} \\
\midrule
\multirow{4}{*}{\makecell{Same \\ Vehicle (S)}} & VRP with Same Vehicle Constraint (VRPSVC)~\citep{kumar2012survey} & \multirow{4}{*}{\makecell{Priority (P)}} & Precedence constrained VRP (PVRP)~\citep{kubo1991precedence} \\
& Clustered VRP (CluVRP)~\citep{battarra2014exact, islam2021hybrid} & & VRP with Relaxed Priority Rules (VRPRP)~\citep{doan2021vehicle} \\
& VRP with Sequential Ordering (VRPSO)~\citep{escudero1988inexact, gambardella2000ant} & & VRP with Multiple Priorities (VRPMP)~\citep{yang2015ant} \\
& VRP with Incompatible Loading Constraint (VRPILC)~\citep{wang2015heuristic} & & VRP with d-Relaxed Priority Rule (VRP-dRP)~\citep{dasari2023two, dos2024formulations} \\
\bottomrule
\end{tabular}%
}
\label{tab:VRP_Variants}
\end{table}

%% file: 5-Experiments.tex
\section{Experiments}
\label{Experiments}

To evaluate the performance of these LLM-based methods in handling different VRP variants, we assess their ability to generate correct programs within our solver.
To analyze the impact of different solvers on the generated programs, we compare our solver with other solvers (e.g., CPLEX, OR-Tools, and Gurobi).
Additionally, we included experiments with other open-source and closed-source LLMs (e.g., DeepSeek V3 and LLaMA 3.1 70B) to explore their impact on program generation.
Finally, we also conduct an ablation study on our proposed ARS.

\begin{table}[!tb]
\centering
\caption{The performance comparison between ARS and seven LLM-based methods on common problems and RoutBench. The best results among these methods are highlighted in grey. 
}
\label{tab:modeling_performance}
\setlength{\tabcolsep}{4pt}
\begin{tabular}{c|cc|cc|cc}
\hline
\multicolumn{1}{c|}{\multirow{3}{*}{\textbf{Methods}}}                     & \multicolumn{2}{c|}{\multirow{2}{*}{\textbf{Common Problems}}} & \multicolumn{4}{c}{\textbf{RoutBench}} \\ 
\cline{4-7}
\multicolumn{1}{l|}{} & \multicolumn{2}{c|}{}                                 & \multicolumn{2}{c|}{\textbf{RoutBench-S}}         & \multicolumn{2}{c}{\textbf{RoutBench-H}}         \\ 
\cline{2-7}
\multicolumn{1}{l|}{} & \textbf{SR $\uparrow$}                       & \textbf{RER} $\downarrow$                     & \textbf{SR $\uparrow$}              & \textbf{RER} $\downarrow$            & \textbf{SR} $\uparrow$              & \textbf{RER} $\downarrow$             \\ \hline
Standard Prompting   & 41.67\%                   & 6.25\%                   & 37.60\%          & 8.20\%          & 11.60\%          & 15.80\%          \\
CoT                  & 37.50\%                   & 8.33\%                   & 37.40\%          & 9.60\%          & 13.40\%          & 14.80\%          \\
Reflexion           & 45.83\%                   & 6.25\%                   & 41.80\%          & 4.60\%          & 15.20\%          & \cellcolor{gray!30}{7.60\%}           \\
PHP                  & 33.33\%                   & 10.42\%                  & 32.60\%          & 12.00\%         & 11.20\%           & 17.40\%          \\
CoE                  & 37.50\%                   & 2.08\%                   & 34.20\%          & 8.00\%          & 11.40\%          & 12.20\%          \\
Self-debug           & 43.75\%                   & \cellcolor{gray!30}{0.00}\%                   & 34.60\%          & \cellcolor{gray!30}{4.00}\%          & 10.80\%          & \cellcolor{gray!30}{7.60\%}           \\
Self-verification    & 43.75\%                   & 2.08\%                   & 34.00\%          & 5.20\%          & 15.60\%          & 9.20\%           \\
ARS (Ours)           & \cellcolor{gray!30}{91.67\%}          &  \cellcolor{gray!30}{0.00\%}          &  \cellcolor{gray!30}{73.20\%} &  5.20\% &  \cellcolor{gray!30}{46.80\%} &  11.80\%
\\ \hline
\end{tabular}
\end{table}

\begin{figure*}[]
    \centering
    \begin{minipage}[]{0.49\textwidth}
        \centering
        \includegraphics[width=1\linewidth]{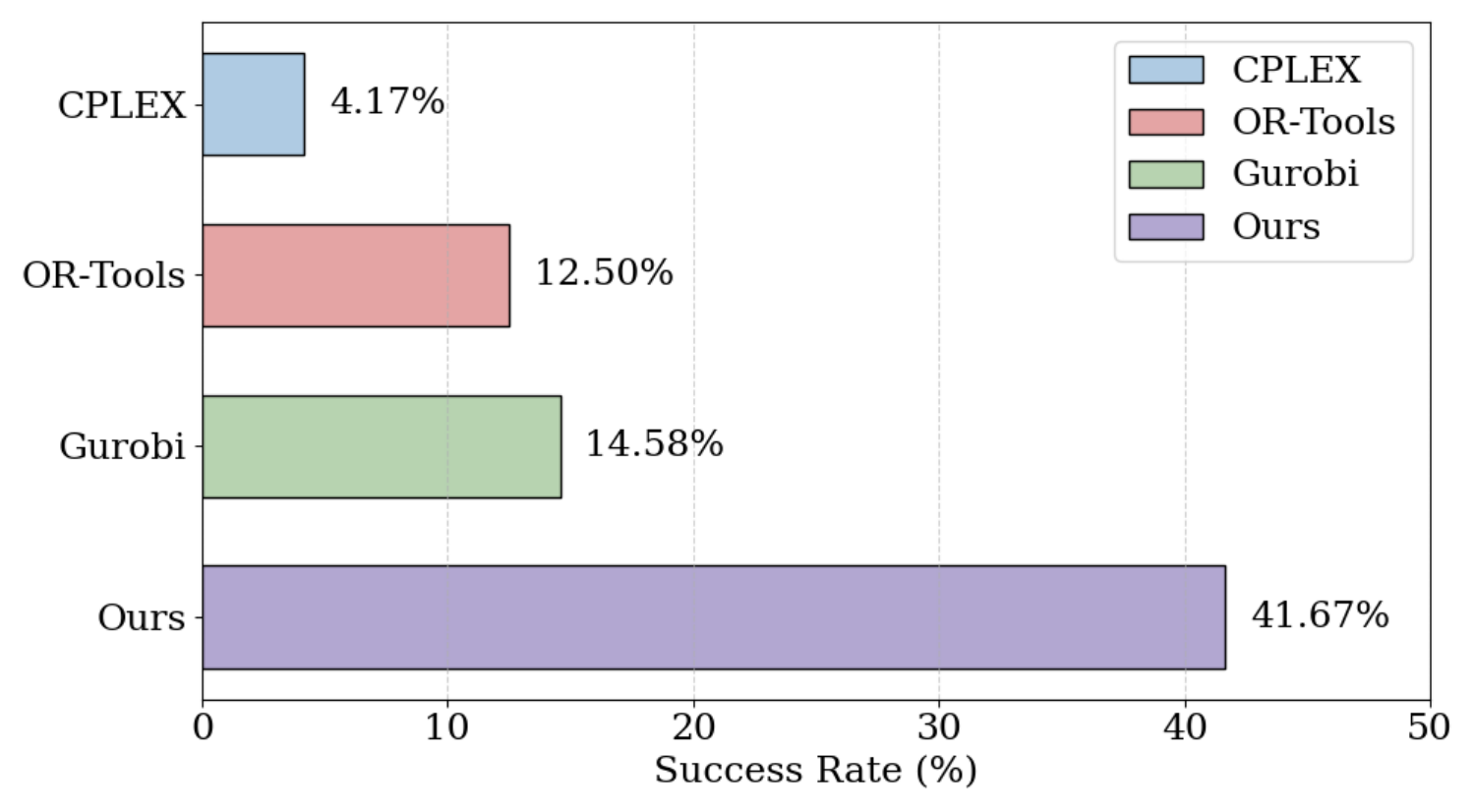}
        \caption{Comparison of the success rate for different solvers with standard prompting.}
        \label{fig:solver}
    \end{minipage}
    \hfill
    \begin{minipage}[]{0.49\textwidth}
        \centering
        \includegraphics[width=0.95\linewidth]{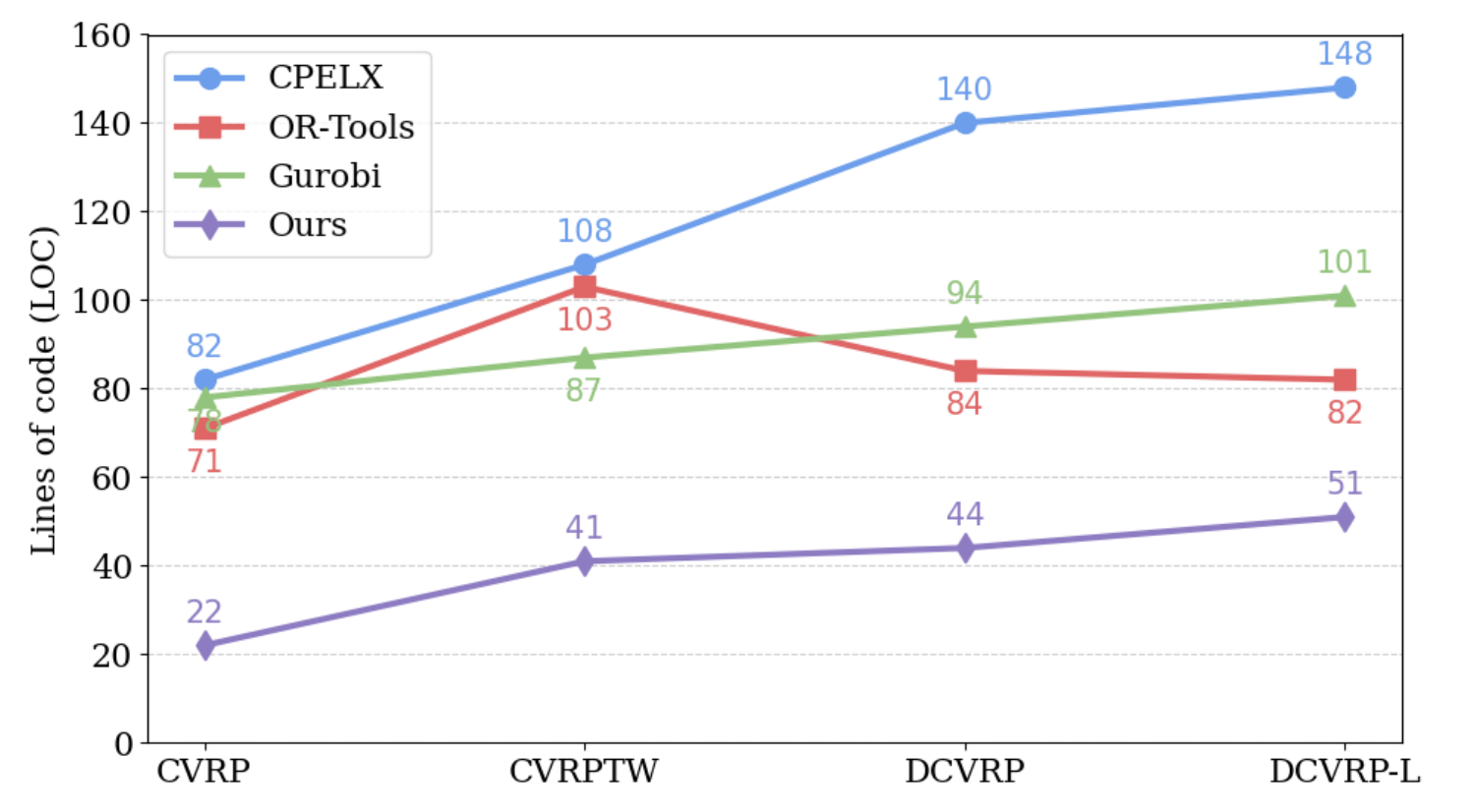}
        \caption{Compares the lines of code generated by LLMs for different solvers on four VRP variants.}
        \label{fig:solver_LOC}
    \end{minipage}
\end{figure*}

\subsection{Comparison with LLM-Based Methods}

The first experiment aims to evaluate the ability of ARS to produce successful programs. We compare our proposed ARS with seven LLM-based methods, including Standard Prompting, Chain of Thought (CoT)~\citep{wei2022chain}, Reflexion~\citep{shinn2024reflexion}, Progressive-Hint Prompting (PHP)~\citep{zheng2023progressive}, Chain-of-Experts (CoE)~\citep{xiao2023chain}, Self-debug~\citep{chen2023teaching}, and Self-verification~\citep{huang2024words}.
To focus on program generation, all methods are implemented using our backbone algorithm framework to handle various VRP variants.

This experiment is conducted on 48 common problems and RoutBench using GPT-4o. We use two metrics to evaluate the results. The Success Rate (SR) measures the proportion of the program where generated solutions pass the validation process. The Runtime Error Rate (RER) quantifies the percentage of programs that fail due to issues such as runtime errors, incorrect API usage, or syntax mistakes.

As shown in Table~\ref{tab:modeling_performance}, ARS significantly outperforms other LLM-based methods in generating correct programs for both common problems and RoutBench. It is evident that these compared methods exhibit unsatisfactory performance across all problems, particularly on the RoutBench-H problems, where their SR is merely around 10\%.
None of the compared algorithms achieved an SR of 50\% or higher.
In contrast, ARS achieved an SR of 91.67\% on the common problems.
Moreover, on RoutBench, ARS generated correct solutions for 60\% of the VRP variants, outperforming all seven other LLM-based methods by at least 30\% in terms of SR.
These results highlight ARS's competitiveness in addressing complex VRP variants.

\subsection{Comparison with Different Solvers}

In the previous experiment, to focus on generating correct programs, all methods used the same solvers to handle VRP variants. To explore the impact of different solving tools on program generation, this experiment uses Standard Prompting with four solvers (CPLEX, OR-Tools, Gurobi, and our solver) for common problems.

As shown in Figures~\ref{fig:solver} and~\ref{fig:solver_LOC}, under the standard prompting, our solver achieves the highest Success 
Rate (SR) on the common problems, and requires the fewest Lines of Code (LOC) compared to other solvers.
This is because, in our framework, the LLM only needs to write constraint codes using straightforward Python syntax, while other solvers are limited by their specific implementations and require highly standardized modeling.
For the same VRP variants, our solving framework makes program generation much simpler compared to these solvers.
Examples of solver codes are provided in Appendix~\ref{Example_codes}.

To evaluate the performance of these solvers on different VRP variants, we conduct a comparative study with other solvers across four VRP variants (e.g., CVRP, CVRPTW, DCVRP, and DCVRP-L).
Details of these problem sets are provided in Appendix~\ref{10-A-Problem_set}.
These solvers usually require considerable time to solve VRPs, so we allocate 25, 50, and 100 minutes for problem sizes of 25, 50, and 100, respectively.
As can be seen from Table~\ref{tab:diff-VRP}, our solver achieved the optimal results in all instances.
Although HGS and LKH-3 can solve common problems quickly and efficiently, our proposed method is designed to tackle a wide range of VRPs, even those it has never encountered before.

\begin{table}[t]
\centering
\caption{Performance analysis of different solvers on four VRP variants. The table presents the gaps compared to the results obtained by ARS. The best results among the four solvers (e.g., CPLEX, OR-Tools, Gurobi, and ARS) are highlighted in grey.}
\label{tab:diff-VRP}
\renewcommand{\arraystretch}{1.4}
\setlength{\tabcolsep}{4pt}
\resizebox{1\textwidth}{!}{%
\begin{tabular}{c|cccccc|cccccc}
\hline
\multicolumn{1}{l}{} & \multicolumn{6}{c|}{\textbf{CVRP}}                                                                             & \multicolumn{6}{c}{\textbf{CVRPTW}}                                                                           \\ \hline
\multirow{2}{*}{Solvers} & \multicolumn{2}{c}{\textbf{N=25}} & \multicolumn{2}{c}{\textbf{N=50 }} & \multicolumn{2}{c|}{\textbf{N=100 }} & \multicolumn{2}{c}{\textbf{N=25 }} & \multicolumn{2}{c}{\textbf{N=50 }} & \multicolumn{2}{c}{\textbf{N=100 }} \\ \cline{2-13}
 & \textbf{Obj. $\downarrow$\ (Gap)}           & \textbf{Time}      & \textbf{Obj. $\downarrow$\ (Gap)}          & \textbf{Time}       & \textbf{Obj. $\downarrow$\ (Gap)}            & \textbf{Time}      & \textbf{Obj. $\downarrow$\ (Gap)}             & \textbf{Time}    & \textbf{Obj. $\downarrow$\ (Gap)}             & \textbf{Time}    & \textbf{Obj. $\downarrow$\ (Gap)}              & \textbf{Time}    \\
 \hline
HGS                  & 186.9 (0.00\%)      & 0.1s         & 358.0 (0.00\%)        & 0.2s         & 817.8 (0.00\%)        & 0.3s        & 190.3 (0.00\%)      & 0.1s         & 361.4 (0.00\%)       & 0.1s        & 826.3 (0.00\%)       & 0.4s         \\
LKH-3                  & 186.9 (0.00\%)      & 0.1s      & 358.0 (0.00\%)        & 0.1s      & 817.8 (0.00\%)        & 1.3s      & 190.3 (0.00\%)      & 0.1s      & 361.4 (0.00\%)       & 0.1s      & 826.3 (0.00\%)       & 4.2s       \\
CPLEX                   & 187.6 (0.37\%)       & 45s      & 362.7 (1.31\%)      & 47m       & 841.6 (2.91\%)        & 1.1h     & \cellcolor{gray!30}{190.3 (0.00\%)}           & 1.3s         & \cellcolor{gray!30}{361.4 (0.00\%)}          & 5.2m         & \cellcolor{gray!30}{826.3 (0.00\%)}           & 1.3h         \\
OR-Tools             & \cellcolor{gray!30}{186.9 (0.00\%)}      & 4.2s       & 358.8 (0.22\%)      & 7.2s       & 849.2 (3.84\%)       & 1.0m       & \cellcolor{gray!30}{190.3 (0.00\%)}      & 0.5s       & 362.5 (0.30\%)       & 9.6s      & 828.1 (0.21\%)       & 3.1m        \\
Gurobi               & \cellcolor{gray!30}{186.9 (0.00\%)}      & 1.8m       & \cellcolor{gray!30}{358.0 (0.00\%)}        & 5.6m       & 828.0 (1.25\%)          & 1.3h      & \cellcolor{gray!30}{190.3 (0.00\%)}      & 10s        & \cellcolor{gray!30}{361.4 (0.00\%)}       & 1.2m      & \cellcolor{gray!30}{826.3 (0.00\%)}       & 1.4h       \\
Ours                 & \cellcolor{gray!30}{186.9 (0.00\%)}      & 3.2s       & \cellcolor{gray!30}{358.0 (0.00\%)}        & 5.4s       & \cellcolor{gray!30}{817.8 (0.00\%)}        & 4.6m      & \cellcolor{gray!30}{190.3 (0.00\%)}      & 5.8s       & \cellcolor{gray!30}{361.4 (0.00\%)}       & 23s       & \cellcolor{gray!30}{826.3 (0.00\%)}       & 1.5m      \\

\hline

\multicolumn{1}{l}{} & \multicolumn{6}{c|}{\textbf{DCVRP}}                                                                            & \multicolumn{6}{c}{\textbf{DCVRP-L}}                                                                          \\ \hline
\multirow{2}{*}{Solvers}            & \multicolumn{2}{c}{\textbf{N=25 }} & \multicolumn{2}{c}{\textbf{N=50 }} & \multicolumn{2}{c|}{\textbf{N=100 }} & \multicolumn{2}{c}{\textbf{N=25 }} & \multicolumn{2}{c}{\textbf{N=50 }} & \multicolumn{2}{c}{\textbf{N=100 }} \\  
\cline{2-13}
              & \textbf{Obj. $\downarrow$\ (Gap)}           & \textbf{Time}      & \textbf{Obj. $\downarrow$\ (Gap)}          & \textbf{Time}       & \textbf{Obj. $\downarrow$\ (Gap)}            & \textbf{Time}      & \textbf{Obj. $\downarrow$\ (Gap)}             & \textbf{Time}    & \textbf{Obj. $\downarrow$\ (Gap)}             & \textbf{Time}    & \textbf{Obj. $\downarrow$\ (Gap)}              & \textbf{Time}    \\
 \hline
 CPLEX                   & 213.3 (9.16\%)       & 35s      & 392.1 (6.98\%)      & 50m       & 878.4 (5.65\%)        & 1.2h     & \cellcolor{gray!30}{215.3 (0.00\%)}           & 1.6m         & 390.7 (1.11\%)          & 48m          & 871.0 (2.68\%)             & 1.6h         \\
OR-Tools             & 253.4 (29.68\%)      & 0.7s      & 426.0 (16.23\%)       & 39s       & 893.6 (7.48\%)        & 1.4h       & 226.3 (5.11\%)         & 1.4s    & 405.8 (5.02\%)         & 1.5m     & 874.6 (3.10\%)          & 11m     \\
Gurobi               & 219.2 (12.18\%)       & 5.2m      & 395.1 (7.80\%)      & 40m        & 874.8 (5.22\%)        & 1.1h      & 219.2 (1.81\%)         & 2.4m    & 395.5 (2.36\%)         & 44m     & 876.8 (3.36\%)          & 1.0h    \\
Ours                 & \cellcolor{gray!30}{195.4 (0.00\%)}       & 6s        & \cellcolor{gray!30}{366.5 (0.00\%)}      & 1.8m       & \cellcolor{gray!30}{831.4 (0.00\%)}        & 18m       & \cellcolor{gray!30}{215.3 (0.00\%)}         & 2.4s    & \cellcolor{gray!30}{386.4 (0.00\%)}         & 26s     & \cellcolor{gray!30}{848.3 (0.00\%)}          & 19m    
  
\\ \hline
\end{tabular}%
}
\end{table}

\subsection{Evaluation with Different LLMs}

We further investigate the impact of using different LLMs for generating correct programs to solve VRP variants under standard prompting and ARS. The LLMs evaluated include GPT-3.5-Turbo, GPT-4o, DeepSeek-V3, and LLaMA-3.1-70B.

\begin{figure*}[t]
    \centering
        \begin{subfigure}[RoutBench-S]{\label{fig:a}\includegraphics[scale=0.123]{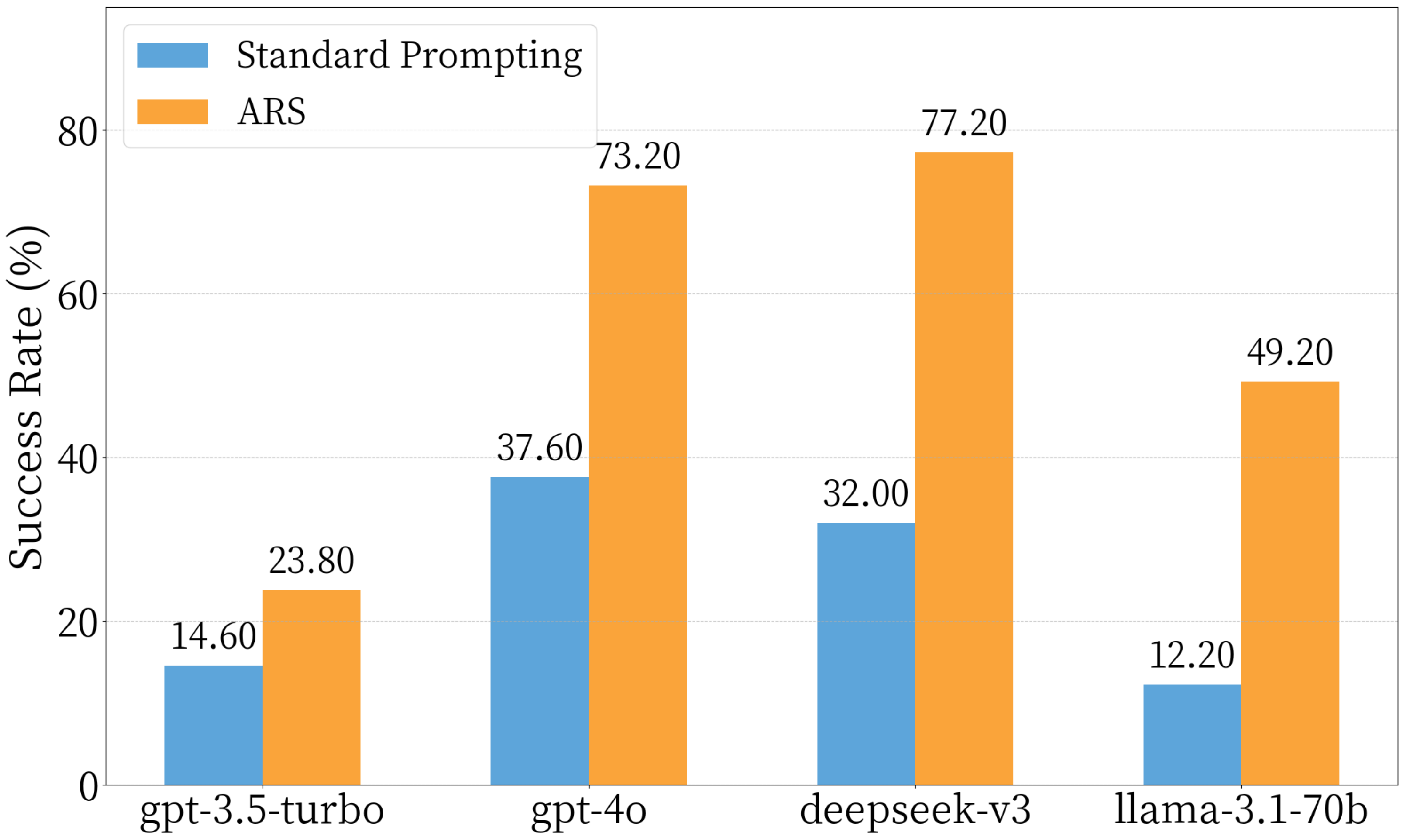}}
        \end{subfigure}
        \hfil
         \begin{subfigure}[RoutBench-H]{\label{fig:b}\includegraphics[scale=0.125]{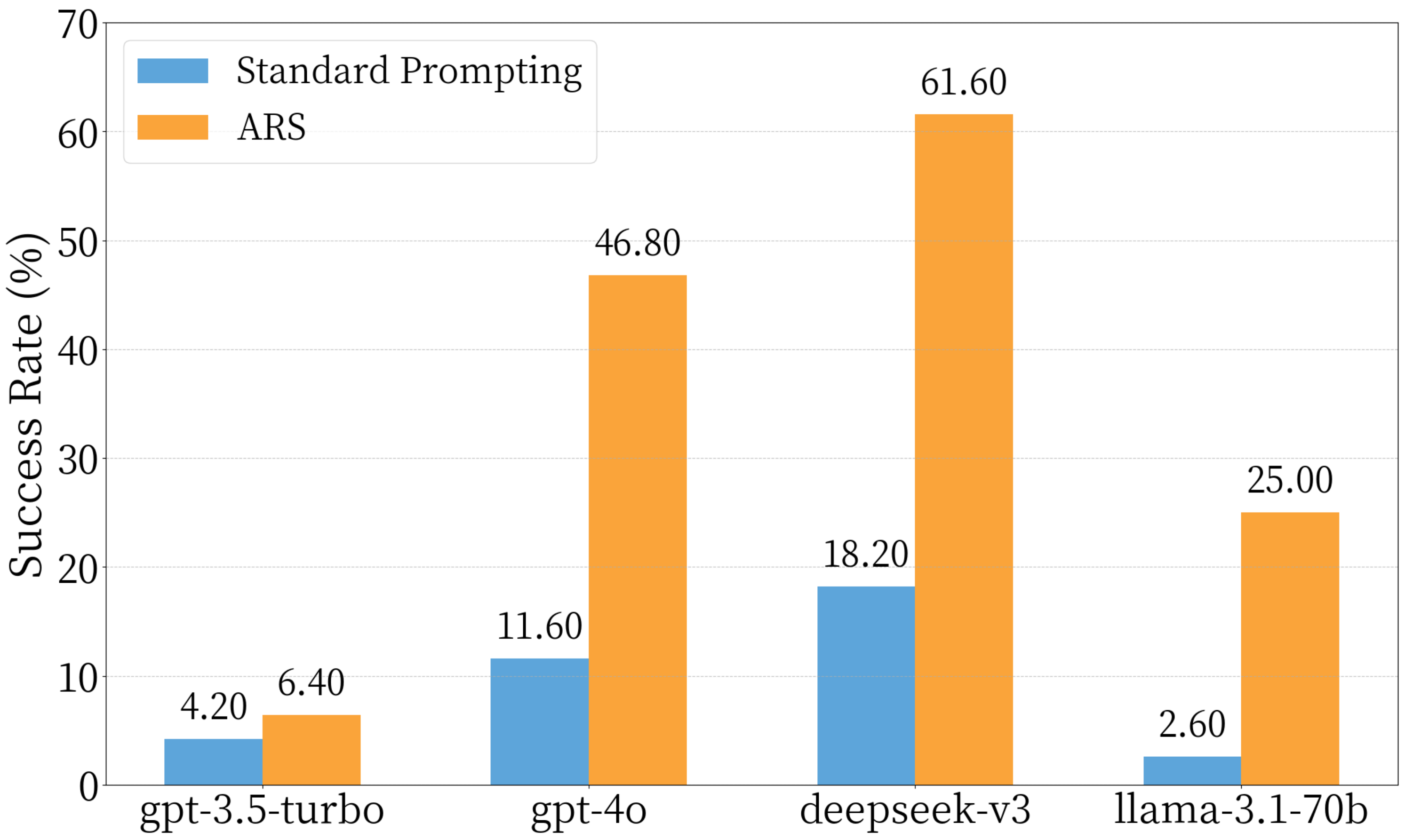}}
         
         \end{subfigure}
        \caption{The performance of standard prompting and ARS is compared across various LLMs. ARS demonstrates compatibility with different models and shows clear advantages in RoutBench.}
        \label{fig:diff_LLMs}
    \end{figure*}

As shown in Figure~\ref{fig:diff_LLMs}, the success rate of solving VRP variants varies across different LLMs. All methods benefit from more advanced LLMs, leading to improved accuracy. The results indicate that DeepSeek-V3 is the most effective LLM for handling VRP variants among these LLMs. When using DeepSeek-V3, ARS achieves an SR of 77.20\% on simple problems in RoutBench. The improvement is even more pronounced for RoutBench-H problems, where ARS attains an SR of 61.60\%.

\begin{table}[!t]
\centering
\caption{Ablation study on the effect of component removal on ARS's ability to generate correct programs. The best results among these methods are highlighted in grey.}
\label{tab:Ablation}
\renewcommand{\arraystretch}{1.18}
\setlength{\tabcolsep}{15pt}
\resizebox{0.6\textwidth}{!}{
\begin{tabular}{ccc}
\hline
\multirow{2}{*}{\textbf{Methods}}      & \multicolumn{2}{c}{\textbf{Common Problems}} \\ \cline{2-3}
                              & \textbf{SR} $\uparrow$              & \textbf{RER} $\downarrow$\\ \hline
ARS (Full)               & \cellcolor{gray!30}{91.67\%}  & \cellcolor{gray!30}{0.00\%} \\
w/o Constraint Selection & 62.50\%           & 2.08\%          \\
w/o Database             & 35.42\%           & 12.50\%   \\ \hline
\end{tabular}}
\end{table}

\subsection{Ablation Study}

An ablation study is conducted to evaluate the program generation component of ARS. To analyze the impact of our database and constraint selection step on ARS, experiments are performed by removing each of these components separately. 

As shown in Table~\ref{tab:Ablation}, removing the constraint selection step leads to a decrease in the SR of ARS.
Without this step, ARS acquires all six representative constraints, which may not be relevant to the current VRP and can mislead the LLM. 
Compared to constraint selection, removing the database has a more significant impact on the performance of ARS in generating correct programs. As previously discussed, ARS can be refined based on relevant constraints, thereby focusing on the distinctive aspects of a VRP variant. However, without reference to relevant constraints, the LLM must independently generate all the constraints, imposing a higher requirement on the LLM.

The results of this ablation study demonstrate that ARS achieves optimal program generation performance only when both components are utilized.

%% file: 6-Conclusion.tex
\section{Conclusion, Limitation, and Future Work}
\label{Conclusion}

\textbf{Conclusion.} In this paper, we have proposed ARS, a framework that leverages LLMs to automatically design constraint-aware heuristics, enhancing a heuristic routing optimization framework for VRP variants with complex and practical constraints.
Moreover, we have developed and presented RoutBench, a comprehensive benchmark consisting of 1,000 VRP variants derived from 24 VRP constraints.
Each variant includes a problem description, instance data, and validation code, enabling standard evaluation of routing solvers. 
The experimental evaluations of ARS have been particularly promising, demonstrating a superior performance over existing LLM-based methods and commonly used solvers.

\textbf{Limitation and Future Work.} 
Designing specialized algorithms for various problems often achieves better solving capability and efficiency.
In this paper, we mainly focus on generating correct solving programs for a wide range of VRPs using a unified heuristic solving framework.
In the future, the solver framework can be enhanced to incorporate LLMs for automated algorithm design tailored to different problems, further improving its solving capability and efficiency.

%% file: 10-Appendix.tex
This is an appendix for “ARS: Automatic Routing Solver with Large Language Models”. Specifically, we provide:

\begin{itemize}
    \item Related works on heuristics, NCO, and LLMs for VRPs (Appendix~\ref{Background}).
    \item Detailed explanation of the methodology, including the prompts used in ARS, examples of LLM outputs, and the operators employed (Appendix~\ref{11-A_Method}).
    \item Details of 48 common problems and constraints for 24 VRP variants (Appendix~\ref{Appendix:Variants}).
    \item More experimental results and analyses including ARS analysis, LLM-suggested constraints, enhancements, and CVRPLib evaluation (Appendix~\ref{Appendix-problems-sets}).
    \item Code examples for our solver, CPLEX, OR-Tools, and Gurobi (Appendix~\ref{Example_codes}).
    \item Function template for the Constraint Scoring Program (Appendix~\ref{12-A-vio_temp}).
    \item Database of six representative constraints and the base case (Appendix~\ref{12-A-Database}).
    \item The potential societal impact of this work (Appendix~\ref{Broader_Impacts}).
    \item The licenses and URLs of the baseline methods (Appendix~\ref{13-Appendix-Baselines+Licenses}).
\end{itemize}

\section{Related works} \label{Background}

The Vehicle Routing Problem (VRP) is a classical combinatorial optimization problem that seeks optimal routes for vehicles to serve customers under constraints like capacity and time windows ~\citep{dantzig1959truck}. Over the years, many VRP variants have been developed and extensively studied, including the Capacitated VRP (CVRP) ~\citep{lysgaard2004new}, the VRP with Time Windows (VRPTW) ~\citep{solomon1987algorithms}, and the Multi-Depot VRP (MDVRP) ~\citep{yuceer1997multi}. However, solving these VRP variants often requires experts to deeply understand the specific problem characteristics, including the constraints, customer demands, and operational rules. This process involves carefully analyzing the problem, designing appropriate models, and implementing customized algorithms. Such methods usually require complex coding and are limited in their ability to address only a small number of VRP variants, making them less flexible and scalable for the diverse and evolving challenges in real-world applications.

\subsection{Heuristics for VRPs}

Traditional methods for solving VRP often rely on heuristics. 
Some simple heuristics, such as the Greedy algorithm and hill-climbing, are commonly used to solve VRP. The Greedy algorithm builds a solution step-by-step by making the most immediate, optimal choice at each step, though it often leads to suboptimal global solutions~\citep{shafique2005noniterative}. Hill-climbing, on the other hand, iteratively improves a solution by moving to a better neighboring solution, but it is prone to getting stuck in local optima~\citep{gent1993towards}. To address these limitations, more advanced metaheuristics have been developed. Simulated Annealing (SA) probabilistically accepts worse solutions during the search process to escape local optima, mimicking the physical annealing process~\citep{kirkpatrick1983optimization}. Tabu Search (TS) enhances local search by using a tabu list to prevent revisiting previously explored solutions, enabling it to explore broader solution spaces~\citep{glover1989tabu}.

In addition to these, state-of-the-art approaches like the Hybrid Genetic Search (HGS) and Lin-Kernighan-Helsgaun (LKH-3) algorithms have achieved remarkable success in solving VRPs. HGS combines genetic algorithms with heuristics tailored for specific types of VRPs, efficiently balancing exploration and exploitation~\citep{vidal2013hybrid}. It is particularly powerful for large-scale and complex VRPs. LKH-3, an extension of the classic Lin-Kernighan heuristic, is highly effective for solving Traveling Salesman Problems (TSP) and TSP-based VRPs, leveraging advanced search strategies and efficient implementations to achieve near-optimal solutions~\citep{helsgaun2017extension}.

Adaptive Large Neighborhood Search (ALNS) represents a more dynamic and flexible approach~\citep{ropke2006adaptive}. It adaptively selects different neighborhood operators based on their performance during the search process, making it highly effective for solving complex VRP variants~\citep{mara2022survey}. Recent advancements in ALNS integrate machine learning techniques to predict the most effective operators and reinforcement learning to optimize selection policies~\citep{wang2024deep}. Hybrid ALNS approaches, such as combining ALNS with branch-and-price methods, further enhance their ability to solve constrained and large-scale VRPs~\citep{vidal2022hybrid}.

Despite their success, these methods often require experts to deeply understand the specific VRP variant, carefully model the problem, and implement customized algorithms. This reliance on expert knowledge limits the scalability of these approaches to handle a broader range of VRP variants without significant manual effort.

\subsection{NCO for VRPs}

Neural combinatorial optimization (NCO) represents a paradigm shift in solving VRPs by leveraging deep learning models to directly learn problem-solving strategies from data.
Several notable NCO approaches have been proposed to address the challenge of solving multiple VRP variants within a unified framework. For instance, MTPOMO tackles cross-problem generalization by representing VRPs as combinations of shared attributes, allowing a single model to solve unseen variants in a zero-shot manner~\citep{liu2024multi}. MVMoE employs a multi-task learning framework with a mixture-of-experts architecture, using a hierarchical gating mechanism to balance model capacity and computational efficiency, achieving strong results across ten unseen VRP variants~\citep{zhou2024mvmoe}. CaDA further advances the field by incorporating a constraint-aware dual-attention mechanism, which effectively captures both global and local problem-specific information, enabling state-of-the-art performance on sixteen VRP variants~\citep{li2024cada}.

However, despite these advancements, current NCO methods still face significant limitations. They typically require manual modifications to adapt algorithms for new VRP variants, limiting their scalability and practicality for real-world applications with highly diverse constraints.

\subsection{LLMs for VRPs}

Recent advancements in large language models (LLMs) have introduced new possibilities for solving vehicle routing problems (VRPs) by leveraging their capacity to encode and process complex optimization tasks~\citep{huang2024words}. 
LLM-based automatic heuristic design (AHD) has emerged as a promising approach, enabling the generation of high-quality heuristics for problems like the traveling salesman problem (TSP) and capacitated VRP (CVRP) without extensive domain expertise. Methods such as Evolutionary Optimization Heuristics (EoH) integrate LLMs with evolutionary computation (EC) to iteratively refine a population of heuristics, facilitating automated discovery of effective solutions~\citep{liu2024evolution}. However, population-based approaches often converge prematurely to local optima. To overcome this, Monte Carlo Tree Search-based AHD (MCTS-AHD) organizes LLM-generated heuristics into a tree structure, enabling deeper exploration of the search space and better utilization of underperforming heuristics~\citep{zheng2025monte}.

Other studies have explored the application of LLMs to different VRP variants, showcasing innovative approaches and promising results. For example, LLM-driven Evolutionary Algorithms (LMEA) utilize LLMs as evolutionary optimizers, achieving competitive results on TSPs with minimal domain knowledge~\citep{liu2024large}. Mechanisms like self-adaptation help balance exploration and exploitation, effectively avoiding local optima. Similarly, an approach proposed by Huang et al.~\citep{huang2024words} enables LLMs to directly generate executable programs for VRPs from natural language task descriptions. This method is further enhanced by a self-reflection framework, which allows LLMs to debug and verify their solutions, significantly improving feasibility, optimality, and efficiency. These early explorations highlight the potential of LLMs in addressing VRPs and advancing the field.

Another line of research explores transforming textual problem descriptions into mathematical formulations and executable code that can be processed by external solvers~\citep{tang2024orlm}. This approach benefits from LLMs’ ability to interpret user queries and generate structured outputs, enabling the automation of optimization tasks. In parallel, multi-agent systems have been introduced to coordinate LLM-based agents for tasks such as problem formulation, programming, and evaluation~\citep{xiao2023chain}. Separately, the DRoC framework introduces a novel method for solving complex VRPs by decomposing constraints, retrieving external knowledge through a retrieval-augmented generation (RAG) approach, and integrating it with the model’s internal knowledge. By dynamically optimizing program generation, this framework has demonstrated significant improvements in both accuracy and runtime efficiency across 48 VRP variants~\citep{jiang2025droc}.
However, despite these innovations, these methods remain inherently constrained by the scope of knowledge encoded within pre-trained models, particularly in generating solver-specific code. This limitation poses significant challenges for LLMs in addressing novel or highly complex problems~\citep{zhang2024solving}.

\input{11-A-Detail_Metho}

\newpage

\section{VRP Variants}\label{Appendix:Variants}

Common problems in VRP are typically constructed based on a set of fundamental constraints, such as vehicle capacity, distance limits, time windows, pickup and delivery, same vehicle, and priority. These problems are widely used to evaluate the performance of multi-task algorithms~\citep{jiang2025droc, liu2024multi, li2024cada}. By combining six representative constraints, excluding cases where vehicle capacity conflicts with pickup and delivery, a subset of 48 common problems is generated, as shown in Table~\ref{table:48_VRPs}.

\input{Tabs/48VRP-variants}

\input{Tabs/24VRP-Prompt}

\newpage

\section{Experimental Details} \label{Appendix-problems-sets}

\textbf{Experimental environment:} Experiments are performed on a computer with an Intel Xeon Gold 6248R Processor (3.00 GHz), 128 GB system memory, and Windows 10.

\subsection{Analysis of ARS in RoutBench}

In RoutBench, the ARS heatmaps illustrate the frequency of simultaneous errors encountered when solving composite Vehicle Routing Problems (VRPs). The horizontal and vertical axes correspond to 24 specific VRPs, with each cell representing the total number of errors occurring when solving a composite problem that includes both the row and column problems. The diagonal values indicate the total number of errors for individual problems, reflecting their inherent difficulty.

\begin{figure*}[!ht]
    \centering
    \includegraphics[width=0.95\linewidth]{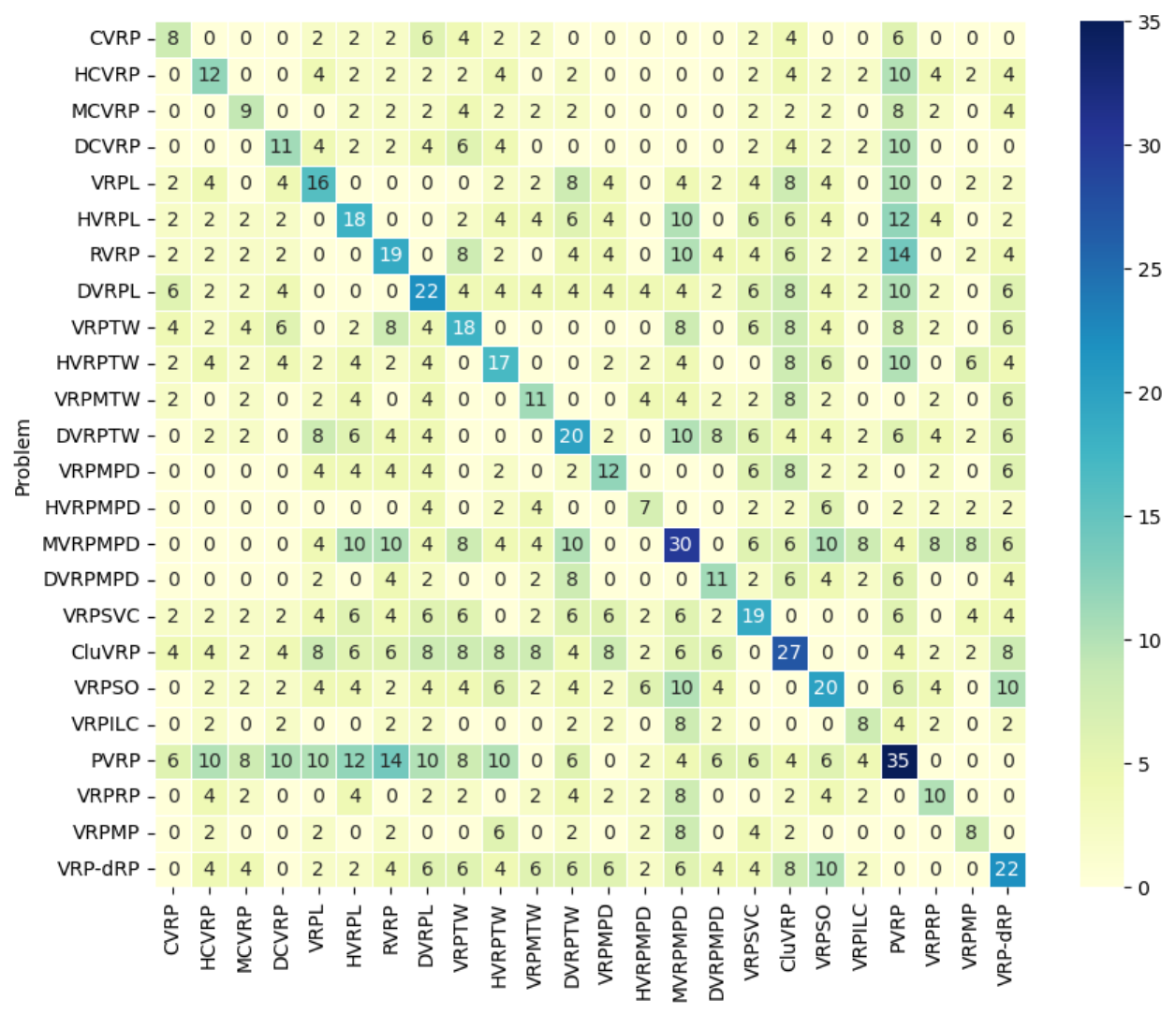}
    \caption{The heatmap of ARS within RoutBench-S shows the number of times errors occur simultaneously for the corresponding row (horizontal axis) and column (vertical axis). The diagonal values represent the number of errors for each individual problem.}
    \label{fig:hot-simple}
\end{figure*}

In the RoutBench-S, the number of errors in priority problems is significantly higher than in other types of problems. This may be due to the inability of LLMs to adequately understand and handle priority issues. When transitioning to the RoutBench-H, the four time-window-related problems exhibit a significantly higher number of errors compared to other problem types. This suggests that time-window problems are inherently more complex. In contrast, regardless of whether the problems are RoutBench-S or RoutBench-H, our algorithm performs exceptionally well in terms of modeling success rates for capacity constraints and return point constraints.

\begin{figure*}[!ht]
    \centering
    \includegraphics[width=0.95\linewidth]{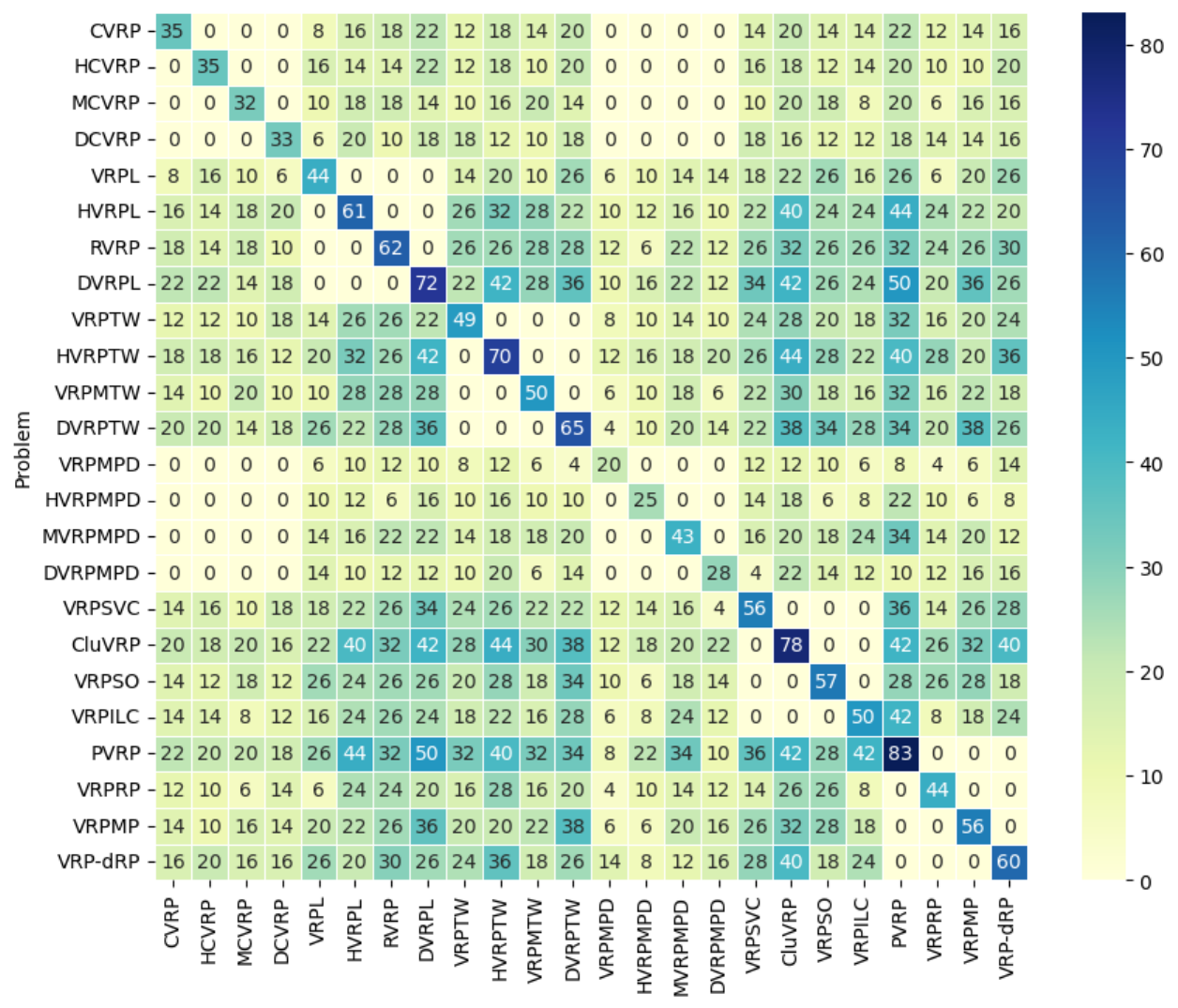}
    \caption{The heatmap of ARS within RoutBench-H shows the number of times errors occur simultaneously for the corresponding row (horizontal axis) and column (vertical axis). The diagonal values represent the number of errors for each individual problem.}
    \label{fig:hot-complex}
\end{figure*}

\subsection{The Number of LLM-Suggested Constraints}

To better understand the relationship between the real constraint count in VRPs and the constraints suggested by the LLM agent, we analyze how the LLM adapts its recommendations based on the problem's complexity.
As shown in Table~\ref{Tab:Num_Constraint}, the data reveals a clear trend where the number of LLM-suggested constraints increases as the real constraint count grows. This suggests that the LLM agent effectively adapts its recommendations based on the complexity of the problem. Interestingly, the LLM agent tends to suggest slightly more constraints than the actual count, likely as a precautionary measure to ensure no potentially relevant constraints are overlooked.

\begin{table}[H]
\centering
\setlength{\tabcolsep}{1mm}
\renewcommand\arraystretch{1}
\caption{Analysis of the relationship between real constraint counts and LLM-suggested constraints in various VRPs, including the total number of problems analyzed for each constraint count and the corresponding average number of suggested constraints by the LLM agent.}
\resizebox{0.9\textwidth}{!}{
\begin{tabular}{ccc}
\toprule
\textbf{Constraint Count} & \textbf{Number of Problems} & \textbf{Average of LLM-Suggested Constraints} \\
\midrule
1 & 8   & 1.25 \\
2 & 92  & 2.43 \\
3 & 400 & 3.81 \\
4 & 263 & 4.84 \\
5 & 237 & 5.63 \\
\bottomrule
\end{tabular}
}
\label{Tab:Num_Constraint}
\end{table}

\subsection{Enhancing ARS with Other Methods}

To further improve the performance of ARS in program generation tasks, we enhanced it with two methods: Reflexion~\citep{shinn2024reflexion} and Self-debug~\citep{chen2023teaching}. Table~\ref{Tab:ARS+other} presents a detailed comparison of the original ARS and its enhanced variants (ARS+Reflexion and ARS+Self-debug) on  RoutBench. The results, obtained using the DeepSeek V3, show that both enhancements lead to improvements in success rate (SR) while simultaneously reducing the error rate (RER). Specifically, ARS+Self-debug achieves the highest success rates and the lowest error rates on RoutBench, demonstrating its effectiveness in refining the program generation process. These findings highlight the potential of incorporating self-improvement mechanisms to boost the performance of ARS.

\begin{table}[hbp]
\centering
\setlength{\tabcolsep}{1mm}
\caption{Performance comparison of ARS and its enhanced variants (ARS+Reflexion and ARS+Self-debug) on RoutBench. The best results among these methods are highlighted in grey.}
\renewcommand{\arraystretch}{1.1}
\setlength{\tabcolsep}{7pt}
\resizebox{0.63\textwidth}{!}{
\begin{tabular}{ccccc}
\toprule
\multirow{2}{*}{\textbf{Methods}} & \multicolumn{2}{c}{\textbf{RoutBench-S}} & \multicolumn{2}{c}{\textbf{RoutBench-H}} \\
\cline{2-5}
 & \textbf{SR $\uparrow$} & \textbf{RER $\downarrow$} & \textbf{SR $\uparrow$} & \textbf{RER $\downarrow$} \\
\midrule
\textbf{ARS+Reflexion}  & 78.20\% & 0.20\% & 63.20\% & 2.00\% \\
\textbf{ARS+Self-debug} & \textbf{78.80\%} & \textbf{0.00\%} & \textbf{68.80\%} & \textbf{0.60\%} \\
\textbf{ARS} & 77.20\% & 2.80\% & 61.60\% & 5.60\% \\
\bottomrule
\end{tabular}} \label{Tab:ARS+other}
\end{table}

\subsection{Performance of ARS across Multiple Runs}

We conducted three independent runs of ARS to evaluate its performance on program generation tasks, with the detailed results presented in Table~\ref{Tab:Three_ARS}. The experiments are performed on RoutBench using the DeepSeek V3, reporting success rates (SR) and error rate reductions (RER) for each trial, along with the averages and standard deviations. The results show consistent performance across the three runs, with low standard deviations in both SR and RER, indicating that ARS performs stably and reliably across different runs and benchmarks.

\begin{table}[hbp]
\centering
\setlength{\tabcolsep}{1mm}
\caption{Detailed results of three independent runs of ARS on RoutBench, including success rates, runtime error rates, averages, and standard deviations to evaluate stability and reliability.}
\renewcommand{\arraystretch}{1.05}
\setlength{\tabcolsep}{7pt}
\resizebox{0.67\textwidth}{!}{
\begin{tabular}{ccccc}
\toprule
\multirow{2}{*}{\textbf{Experiment}} & \multicolumn{2}{c}{\textbf{RoutBench-S}} & \multicolumn{2}{c}{\textbf{RoutBench-H}} \\
\cline{2-5}
 & \textbf{SR $\uparrow$} & \textbf{RER $\downarrow$} & \textbf{SR $\uparrow$} & \textbf{RER $\downarrow$} \\
\midrule
Trial 1 & 77.20\% & 2.80\% & 61.60\% & 5.60\% \\
Trial 2 & 80.60\% & 2.00\% & 64.20\% & 2.60\% \\
Trial 3 & 76.40\% & 2.80\% & 64.80\% & 5.80\% \\
\midrule
Average & \textbf{78.07\%} & \textbf{2.53\%} & \textbf{63.53\%} & \textbf{4.67\%} \\
Standard Deviation & 3.56\% & 0.14\% & 2.14\% & 2.89\% \\
\bottomrule
\end{tabular}} 
\label{Tab:Three_ARS}
\end{table}

\subsection{Best-Known Solutions for RoutBench}\label{10-A-BKS}

The best-known solutions (BKS) can be accessed through the provided link to the RoutBench repository. We provide the BKS for all instances in RoutBench, a comprehensive benchmark that encompasses 1,000 VRP variants with varying problem sizes (25, 50, and 100 nodes). For each instance, the BKS is obtained using ARS, which applies the correct Constraint-Aware Heuristic to ensure feasibility and solution quality. The algorithm is executed under rigorous stopping criteria: it terminates when no improvement is observed for 1,000 consecutive generations or when the runtime exceeds one hour.

\subsection{Problem Set for Test Instances} \label{10-A-Problem_set}

As shown in Table~\ref{Tab:problem_set}, two common problems (e.g., CVRP and CVRPTW) and two dynamic problems (e.g., DCVRP and DCVRP-L) are presented, along with their respective settings and constraints. In this paper, "N" represents the number of points, "C" represents the vehicle capacity, and "L" represents the length of the maximum travel distance.

\begin{table}[H]
\centering
\setlength{\tabcolsep}{1mm}
\caption{Detailed problem set and constraints for test instances, including common and dynamic VRP variants.}
\renewcommand{\arraystretch}{1.4}
\setlength{\tabcolsep}{7pt}
\resizebox{1\textwidth}{!}{
\begin{tabular}{l|l|p{13cm}}
\hline
\textbf{Problem} & \textbf{Problem Set} & \textbf{Problem Description} \\ \hline
CVRP & C=200 & The total load of each route must not exceed the vehicle capacity. \\ \hline
CVRPTW & C=200 & The total load of each route must not exceed the vehicle capacity. The arrival time at each node must meet its specified time window. \\ \hline
DCVRP & C=200 & The total load of each route must not exceed the vehicle capacity. Specifically, for node [19], its base demand is augmented by 5 times the square root of the accumulated travel distance from the depot [0] to that node. \\ \hline
DCVRP-L & C=200, L=150 & The total load of each route must not exceed the vehicle capacity. Specifically, for node [19], its base demand is augmented by 5 times the square root of the accumulated travel distance from the depot [0] to that node. Each route must not exceed 150 units in length. \\ 
\bottomrule
\end{tabular}} 
\label{Tab:problem_set}
\end{table}

\subsection{Evaluation of ARS Performance on CVRPLib}

To further validate the effectiveness of ARS in solving real-world instances, we conducted experiments using five test suites from the CVRPLib\footnote{http://vrp.atd-lab.inf.puc-rio.br/} benchmark datasets. These datasets consist of 99 instances from Sets A, B, F, P, and X~\citep{uchoa2017new}, encompassing graph scales ranging from 16 to 200 nodes, diverse node distributions, and varying customer demands.

In this subsection, we provided benchmark results for state-of-the-art algorithms HGS and LKH-3, as well as some simple heuristics, including the Greedy Algorithm and Hill Climbing, for reference purposes. Additionally, we compared ARS with three widely-used solvers: CPLEX, OR-Tools, and Gurobi.

As shown in Table~\ref{Tab:cvrplib}, ARS achieves the best performance among the four solvers, demonstrating its effectiveness and reliability in solving these challenging instances. The data for HGS, LKH-3, and simple heuristics serve as a reference to contextualize the problem complexity and solution quality.


\begin{table}[H]
\centering
\setlength{\tabcolsep}{1mm}
\caption{Performance comparison of ARS with other methods on CVRPLib. The best results among the four solvers (e.g., CPLEX, OR-Tools, Gurobi, and ARS) are highlighted in grey.}
\renewcommand{\arraystretch}{1.05}
\setlength{\tabcolsep}{10pt}
\resizebox{1\textwidth}{!}{
\begin{tabular}{cc|cccccccc}
\toprule
      &         & \multicolumn{2}{c}{HGS}   & \multicolumn{2}{c}{LKH-3}    & \multicolumn{2}{c}{Greedy} & \multicolumn{2}{c}{Hill Climbing} \\
      & Opt.    & Obj.         & Gap        & Obj.           & Gap         & Obj.              & Gap              & Obj.            & Gap             \\ \midrule
Set A & 1041.9  & 1042.2       & 0.00\%     & 1041.9         & 0.00\%      & 1379.6            & 32.41\%          & 1183            & 13.54\%         \\
Set B & 963.7   & 964.5        & 0.00\%     & 963.7          & 0.00\%      & 1272              & 31.99\%          & 1140.6          & 18.36\%         \\
Set F & 707.7   & 709          & 0.00\%     & 707.7          & 0.00\%      & 1021              & 44.27\%          & 1044.3          & 47.56\%         \\
Set P & 587.4   & 586.9       & 0.00\%     & 587.4          & 0.00\%      & 728.7             & 24.06\%          & 681             & 15.93\%         \\
Set X & 27220.1 & 27223.7      & 0.01\%     & 27281.4        & 0.02\%      & 34254.3           & 25.84\%          & 32301.3         & 18.67\%         \\\midrule
Avg.  & 6104.2  & 6105.3     & 0.00\%     & 6116.4         & 0.00\%      & 7731.1           & 31.71\%          & 7270.1         & 22.81\%         \\\midrule
      &         & \multicolumn{2}{c}{CPLEX} & \multicolumn{2}{c}{OR-Tools} & \multicolumn{2}{c}{Gurobi}           & \multicolumn{2}{c}{ARS}           \\
      & Opt.    & Obj.         & Gap        & Obj.           & Gap         & Obj.              & Gap              & Obj.            & Gap             \\\midrule
Set A & 1041.9  & 1096.5      & 5.24\%     & 1058.9         & 1.63\%      & 1067.3            & 2.44\%           & \textbf{1055.5}          & \textbf{1.31\%}          \\
Set B & 963.7   & 1003.6       & 4.14\%     & 973.3          & 1.00\%      & 990.9             & 2.82\%           & \textbf{973}             & \textbf{0.96\% }         \\
Set F & 707.7   & 789.3        & 11.53\%    & 728.7          & 2.97\%      & 728.7             & 2.97\%           & \textbf{727}             & \textbf{2.73\% }         \\
Set P & 587.4   & 612.5        & 4.27\%     & 592            & 0.78\%      & 594.9             & 1.28\%           & \textbf{591.1}           & \textbf{0.62\% }         \\
Set X & 27220.1 & 32044.1      & 17.72\%    & 28209.6        & 3.64\%      & 28977.7           & 6.46\%           & \textbf{28142.4}         & \textbf{3.39\% }         \\ \midrule
Avg.  & 6104.2  & 7109.2     & 8.58\%     & 6312.5         & 2.00\%      & 6471.9            & 3.19\%           & \textbf{6297.8 }         & \textbf{1.80\% }  \\
\bottomrule
\end{tabular}} 
\label{Tab:cvrplib}
\end{table}

\newpage
\section{Examples of Solver Codes} \label{Example_codes}

To better analyze different methods for solving VRPs, we provide code examples for four approaches: our solver, Gurobi, OR-Tools, and CPLEX. As a case study, these methods are applied to the Capacitated Vehicle Routing Problem with Time Windows (CVRPTW) to illustrate their respective requirements and complexities.

\textbf{Our Solver Code:}

\customcode[1]{Base-model/solver-result-CVRPTW/Our_solver.tex}

\newpage

\textbf{Gurobi Code:}

\customcode[1]{Base-model/solver-result-CVRPTW/Gurobi1.tex}

\customcode[47]{Base-model/solver-result-CVRPTW/Gurobi2.tex}

From the provided code examples, it is clear that our solver requires the least amount of code to be generated by an LLM. This is attributed to the simplicity and flexibility of our method, which avoids the need for verbose or overly rigid programming constructs. In contrast, the code examples for Gurobi, OR-Tools, and CPLEX are more complex, primarily due to their reliance on strict syntax rules and detailed configurations.

\newpage

\textbf{OR-Tools Code:}

\customcode[1]{Base-model/solver-result-CVRPTW/OR-tools1.tex}

\customcode[50]{Base-model/solver-result-CVRPTW/OR-tools2.tex}

\newpage

\textbf{CPLEX Code:}

\customcode[1]{Base-model/solver-result-CVRPTW/CPLEX1.tex}

\customcode[51]{Base-model/solver-result-CVRPTW/CPLEX2.tex}

\input{12-A-Base-models}

\newpage

\section{Potential Societal Impact} \label{Broader_Impacts}

Automatic Routing Solver (ARS) has the potential to drive significant societal advancements by addressing complex real-world Vehicle Routing Problems (VRPs) in industries such as logistics, transportation, and healthcare. ARS leverages Large Language Model (LLM) agents to automate the design of constraint-aware heuristic solvers, offering flexibility and efficiency in solving diverse VRP scenarios.

ARS brings two main strengths to VRP solutions: 1) it dynamically adapts to diverse practical constraints, providing robust solutions without requiring extensive manual design, and 2) it introduces interpretability, enabling decision-makers to better understand and customize routing solutions for specific needs. These features make ARS a valuable tool for optimizing operations, reducing costs, and improving resource utilization across sectors. Additionally, the RoutBench benchmark ensures rigorous evaluation of ARS, further validating its real-world applicability.

However, ARS is not without challenges. Over-reliance on automated solvers may limit human oversight, and misinterpretation of results could lead to suboptimal decisions. Furthermore, the use of LLMs in ARS raises concerns about data security, as sensitive operational or constraint information could inadvertently be exposed. Addressing these risks will be crucial to ensuring ARS's safe and ethical deployment. Lastly, while ARS demonstrates strong performance, its success rate may vary depending on the complexity of constraints, potentially delaying decision-making in highly intricate scenarios.

\input{13-A-Baselines_License}

%% file: 11-A-Detail_Metho.tex
\section{Detailed Methodology} \label{11-A_Method}

\subsection{Prompts of ARS}

Automatic Routing Solver (ARS) is designed to address each VRP variant by leveraging LLMs in three key steps: \textbf{Constraint Selection}, \textbf{Constraint Checking Program Generation}, and \textbf{Constraint Scoring Program Generation}, with a total of three LLM calls. In this section, we describe the prompt engineering involved in each step. These prompts are constructed based on user inputs and several representative constraints stored in the database to generate the \textit{Constraint Checking Program} and the \textit{Constraint Scoring Program}. Variable information, such as user inputs and constraint descriptions, is highlighted in \textcolor{blue!65!black}{blue} for clarity.

\textbf{Step 1: Constraint Selection.} 
In the first step, the ARS identifies constraints relevant to the user's input from the database. This step processes the user input and matches it against the constraints stored in the database. If relevant constraints are found, they are selected for further processing. Otherwise, LLM outputs \textit{"No Relevant Constraint"}. This step ensures that only the constraints relevant to the problem description are considered in subsequent steps.

\begin{dialogbox}[Prompt for Constraint Selection]

\textcolor{red!40!black}{For the description in the VRP problem, identify and provide the relevant constraint types from the following list:}

\textcolor{blue!65!black}{\{constraint\_description\}}\\

\textcolor{red!40!black}{According to the user input:}

\textcolor{blue!65!black}{\{user\_input\}}\\
\textcolor{red!40!black}{If no constraint types match the user input, respond with:} \textcolor{green!30!black}{"No Relevant Constraint"}.\\

Do not give additional explanations.
\end{dialogbox}

\textbf{Step 2: Constraint Checking Program Generation.} 
Based on the relevant constraints selected in the previous step, the ARS uses the LLM to generate a new Constraint Checking Program by taking the selected constraints as references. Specifically, the selected constraints are provided as input to the LLM, which then generates the new program tailored to the problem description and the referenced constraint information.

\begin{dialogbox}[Prompt for Constraint Checking Program Generation]
\textcolor{red!40!black}{As a Python algorithm expert, please implement a function to check the constraints for the vehicle routing problem (VRP) based on the provided description and relevant code.}\\

\textcolor{red!40!black}{User input:}

\textcolor{blue!65!black}{\{user\_input\}}\\

\textcolor{red!40!black}{Relevant Examples:}

\textcolor{blue!65!black}{\{related\_constraints\_and\_codes\}}\\

Do not give additional explanations.
\end{dialogbox}

\textbf{Step 3: Constraint Scoring Program Generation.} 
In the final step, the ARS generates a \textit{Constraint Scoring Program} using the \textit{Constraint Checking Program} developed in the previous step as a foundation. This scoring program evaluates the degree to which the constraints are satisfied by assigning a quantitative score based on the results of the constraint checks.

\begin{dialogbox}[Prompt for Constraint Scoring Program Generation]
\textcolor{red!40!black}{As a Python algorithm expert, please implement a function to calculate the constraint violation score for the Vehicle Routing Problem (VRP) based on the given constraints.}\\

\textcolor{red!40!black}{Function Template:}

\textcolor{blue!65!black}{\{function\_template\}}\\

\textcolor{red!40!black}{Constraints Description:}

\textcolor{blue!65!black}{\{constraints\_description\}}\\

\textcolor{red!40!black}{Check Constraints Code:}

\textcolor{blue!65!black}{\{related\_constraints\_and\_codes\}}\\

Do not give additional explanations.
\end{dialogbox}

\subsection{Examples of LLM Outputs}

This subsection shows an example of using LLM with ARS to solve the CVRP with Incompatible Loading Constraints (CVRP-ILS).
The process includes three LLM calls. The first call selects constraints related to CVRP-ILS. The second call creates a program to check these constraints. The third call creates a program to calculate violation scores.

\begin{dialogbox1}[Example of Step 1: Constraint Selection]

\textbf{First Call Input:}\\
\textcolor{red!40!black}{For the description in the VRP problem, identify and provide the relevant constraint types from the following list:}
\\
1. No Relevant Constraint\\
Example: No relevant constraint. \\
2. Vehicle Capacity Constraint\\
Example: The total load of each route must not exceed the vehicle capacity. \\
3. Distance Limit Constraint\\
Example: Each route's total length must not exceed 100 units.\\ 
4. Time Windows Constraint\\
Example: The arrival time at the node must meet the time windows. \\
5. Pickup and Delivery Constraint\\
Example: At Node [24], 10 units of goods are picked up. Unlike delivery, visiting a pickup node reduces the vehicle's available capacity.\\
6. Same Vehicle Constraint\\
Example: Nodes [1, 10] must be served by the same vehicle. 
\\
7. Priority Constraint\\
Example: Nodes [17, 19] are priority points. \\

\textcolor{red!40!black}{According to the user input:}\\
\textcolor{blue!65!black}{The total load of each route must not exceed the vehicle capacity. Nodes [7, 8] must not be on the same route.}\\
\textcolor{red!40!black}{If no constraint types match the user input, respond with:} \textcolor{green!30!black}{"No Relevant Constraint"}.\\

Do not give additional explanations.\\

\textbf{First Call Output:}\\
-------------------------- Relevant Constraints Type --------------------------\\
1. Constraint type: Vehicle Capacity Constraint\\
2. Constraint type: Same Vehicle Constraint
\end{dialogbox1}

The problem description for CVRP-ILS is: The total load of each route must not exceed the vehicle capacity. Nodes [7, 8] must not be on the same route.

In the first step, the relevant constraints for CVRP-ILS are identified and selected. Therefore, the set of constraints for CVRP-ILS is $S = \{\text{"Vehicle Capacity Constraint"}, \text{"Same Vehicle Constraint"}\}$. These constraints will guide the next step in helping ARS generate constraint checking program for CVRP-ILS.

\begin{dialogbox1}[Example of Step 2: Constraint Checking Program Generation]

\textbf{Second Call Input:}\\
\textcolor{red!40!black}{As a Python algorithm expert, please implement a function to check the constraints for the vehicle routing problem (VRP) based on the provided description and relevant code.}\\

\textcolor{red!40!black}{User input:}\\
\textcolor{blue!65!black}{The total load of each route must not exceed the vehicle capacity. Nodes [7, 8] must not be on the same route.}\\

\textcolor{red!40!black}{Relevant Examples:}

\textcolor{gray}{\# See Appendices~\ref{12-A-Exp-C} and~\ref{12-A-Exp-S}.}\\

Do not give additional explanations.\\

\textbf{Second Call Output:}
\begin{lstlisting}
def check_constraints(solution: VrpState) -> bool:
  """
  Check if the solution satisfies the constraints.

  Args:
    solution (VrpState): An object representing the VRP solution.
      - problem_data (dict): A dictionary with problem details:
        - "edge_weight": A 2D NumPy array of node distances .
        - "demand": A 1D NumPy array of node demands (0 for the depot).
        - "capacity": Maximum load capacity per vehicle.

  Return:
    bool: True if the solution satisfies the constraints, False otherwise.

  Note:
    The above "Args" cannot be added or modified, and no data outside the "solution" object should be used or added.
  """
  # Check vehicle capacity constraint
  for route in solution.routes:
    total_demand = sum(solution.problem_data["demand"][node] for node in route)
    if total_demand > solution.problem_data["capacity"]:
      return False
  
  # Check nodes [7, 8] not on same route constraint
  for route in solution.routes:
    if 7 in route and 8 in route:
      return False
  
  return True
\end{lstlisting}
\end{dialogbox1}

\begin{dialogbox1}[Example of Step 3: Constraint Scoring Program Generation]

\textbf{Third Call Input:}\\
\textcolor{red!40!black}{As a Python algorithm expert, please implement a function to calculate the constraint violation score for the Vehicle Routing Problem (VRP) based on the given constraints.}\\

\textcolor{red!40!black}{Function Template:}

\textcolor{gray}{\# See Appendix~\ref{12-A-vio_temp}.}\\


\textcolor{red!40!black}{Constraints Description:}

\textcolor{blue!65!black}{The total load of each route must not exceed the vehicle capacity. Nodes [7, 8] must not be on the same route.}\\

\textcolor{red!40!black}{Check Constraints Code:}

\textcolor{gray}{\# Use the constraint checking code generated in Step 2.}

\textcolor{blue!65!black}{}

Do not give additional explanations.\\

\textbf{Third Call Output:}
\begin{lstlisting}
def calculate_violation_score(solution: VrpState) -> float:
  """
  Args:
    solution (VrpState): An object representing the VRP solution.
      - problem_data (dict): A dictionary with problem details; Use only defined values in "Check Constraints Code".
      - routes (list): A list of routes, where each route is a list of node IDs (integers, excluding the depot).
      - cost (Callable[[], float]): Implemented method to compute and return the total cost of the solution's routes.
  Return:
    violation_score (float): The calculated constraint violation score, where higher scores indicate more severe violations.

  Note:
    Do not use data or methods that is not contained in the "solution" object.
  """
  violation_score = 0.0
  
  # Check vehicle capacity constraint
  for route in solution.routes:
    total_demand = sum(solution.problem_data["demand"][node] for node in route)
    if total_demand > solution.problem_data["capacity"]:
      violation_score += (total_demand - solution.problem_data["capacity"])
  
  # Check nodes [7, 8] not on same route constraint
  for route in solution.routes:
    if 7 in route and 8 in route:
      violation_score += 1.0
  
  return violation_score
\end{lstlisting}
\end{dialogbox1}

\newpage
\subsection{Operator}\label{Appendix:Oper}

Local search operators are essential components of heuristic and metaheuristic methods, designed to explore the neighborhood of a solution and iteratively improve its quality. These operators are the building blocks for efficiently navigating the search space, balancing exploration and exploitation.
The commonly used operators are as follows:

\textbf{2-opt Operator~\citep{lin1965computer}.} The 2-opt operator is a classical approach originally developed for the Traveling Salesman Problem (TSP). It works by removing two non-adjacent edges in the solution and reconnecting them in a different way, thereby altering the order of nodes. If the new configuration reduces the total cost, it is accepted as an improved solution.

\textbf{Swap Operator~\citep{osman1993metastrategy}.} The Swap operator is another simple yet powerful tool in local search methods. It works by exchanging the positions of two elements within the solution. This operation generates a new configuration, which can help in escaping local optima and promoting diversity in the solution space.

\textbf{Shift Operator~\citep{rosenkrantz1977analysis}.} The Shift operator involves moving an element from one position in the solution to another. This operation changes the relative ordering of elements, redistributing their positions to explore alternative configurations. By shifting elements, the algorithm can adjust the structure of the solution in a more targeted manner, allowing it to overcome local optimality and discover new regions of the solution space.

\textbf{Destroy Operator.} The Destroy operator partially disrupts the current solution by selectively removing some elements. This disruption breaks the local optimality of the solution, allowing for the exploration of new regions in the search space. There are two common implementations of this operator: \textbf{Random Removal} and \textbf{String Removal}.

\begin{itemize}
    \item \textbf{Random Removal}: This method involves removing elements uniformly at random, without relying on specific heuristics or problem-dependent strategies, making it a straightforward yet highly effective approach to diversify the search process and introduce variability into the solution space.
    
    \item \textbf{String Removal}: This method targets sequences of consecutive or related elements (strings), such as partial routes or groups of customers~\cite{christiaens2020slack}. It begins by randomly selecting a "center" customer and removing a string of nearby customers from the route. The string size is randomly determined, constrained by the average route size and a predefined maximum. If constraints on the number of disrupted routes or previously disrupted routes are met, further removal is skipped.
\end{itemize}

\textbf{Repair Operator.} The Repair operator complements the Destroy operator by reinserting removed elements to reconstruct a complete solution, guided by optimization objectives. This combination of destruction and repair allows the algorithm to iteratively refine solutions while maintaining the flexibility to explore new possibilities. One commonly used implementation of the Repair operator is \textbf{Greedy Repair}:

\begin{itemize}
    \item \textbf{Greedy Repair}: This method reinserts removed elements one by one, selecting at each step the position that minimizes the objective function. By considering constraint-aware heuristics during the reinsertion process, it ensures that each step improves solution quality while adhering to problem-specific constraints, effectively balancing optimality and feasibility throughout the search process.
\end{itemize}

In summary, the local search operators discussed above, including 2-opt, Swap, Shift, Destroy, and Repair, play a crucial role in the design of heuristic and metaheuristic algorithms. These operators enable targeted adjustments to the solution, facilitating efficient exploration and exploitation of the solution space. By combining these operators, algorithms can effectively escape local optima and converge toward high-quality solutions.

In our approach, we adopt these efficient operators within a Backbone heuristic framework, which provides the structural foundation for solving complex optimization problems. The framework leverages these operators to iteratively refine solutions, balancing between intensification and diversification.

%% file: Tabs/48VRP-variants.tex

\begin{table}[!h]
\centering

\caption{The 48 common problems constructed by six representative constraints.}~\label{table:48_VRPs}
\tiny
\resizebox{0.95\textwidth}{!}{%
\begin{tabular}{l|cccccc}
\hline \rule{0pt}{2.5mm}
          & Vehicle & Distance& Time& Pickup and& Same & Priority\\
        &Capacity  &Limit  &Windows  &Delivery& Vehicle &  \\ [0.5mm]
\hline
VRP         &   &   &   &   &   &   \\
PVRP        &   &   &   &   &   & $\surd$ \\
VRPS        &   &   &   &   & $\surd$ &   \\
PVRPS       &   &   &   &   & $\surd$ & $\surd$ \\
VRPPD      &   &   &   & $\surd$ &   &   \\
PVRPPD     &   &   &   & $\surd$ &   & $\surd$ \\
VRPPDS     &   &   &   & $\surd$ & $\surd$ &   \\
PVRPPDS    &   &   &   & $\surd$ & $\surd$ & $\surd$ \\
VRPTW       &   &   & $\surd$ &   &   &   \\
PVRPTW      &   &   & $\surd$ &   &   & $\surd$ \\
VRPSTW      &   &   & $\surd$ &   & $\surd$ &   \\
PVRPSTW     &   &   & $\surd$ &   & $\surd$ & $\surd$ \\
VRPPDTW    &   &   & $\surd$ & $\surd$ &   &   \\
PVRPPDTW   &   &   & $\surd$ & $\surd$ &   & $\surd$ \\
VRPPDSTW   &   &   & $\surd$ & $\surd$ & $\surd$ &   \\
PVRPPDSTW  &   &   & $\surd$ & $\surd$ & $\surd$ & $\surd$ \\
VRPL        &   & $\surd$ &   &   &   &   \\
PVRPL       &   & $\surd$ &   &   &   & $\surd$ \\
VRPLS       &   & $\surd$ &   &   & $\surd$ &   \\
PVRPLS      &   & $\surd$ &   &   & $\surd$ & $\surd$ \\
VRPPDL     &   & $\surd$ &   & $\surd$ &   &   \\
PVRPPDL    &   & $\surd$ &   & $\surd$ &   & $\surd$ \\
VRPPDLS    &   & $\surd$ &   & $\surd$ & $\surd$ &   \\
PVRPPDLS   &   & $\surd$ &   & $\surd$ & $\surd$ & $\surd$ \\
VRPLTW      &   & $\surd$ & $\surd$ &   &   &   \\
PVRPLTW     &   & $\surd$ & $\surd$ &   &   & $\surd$ \\
VRPLSTW     &   & $\surd$ & $\surd$ &   & $\surd$ &   \\
PVRPLSTW    &   & $\surd$ & $\surd$ &   & $\surd$ & $\surd$ \\
VRPPDLTW   &   & $\surd$ & $\surd$ & $\surd$ &   &   \\
PVRPPDLTW  &   & $\surd$ & $\surd$ & $\surd$ &   & $\surd$ \\
VRPPDLSTW  &   & $\surd$ & $\surd$ & $\surd$ & $\surd$ &   \\
PVRPPDLSTW &   & $\surd$ & $\surd$ & $\surd$ & $\surd$ & $\surd$ \\
CVRP        & $\surd$ &   &   &   &   &   \\
PCVRP       & $\surd$ &   &   &   &   & $\surd$ \\
CVRPS       & $\surd$ &   &   &   & $\surd$ &   \\
PCVRPS      & $\surd$ &   &   &   & $\surd$ & $\surd$ \\
CVRPTW      & $\surd$ &   & $\surd$ &   &   &   \\
PCVRPTW     & $\surd$ &   & $\surd$ &   &   & $\surd$ \\
CVRPSTW     & $\surd$ &   & $\surd$ &   & $\surd$ &   \\
PCVRPSTW    & $\surd$ &   & $\surd$ &   & $\surd$ & $\surd$ \\
CVRPL       & $\surd$ & $\surd$ &   &   &   &   \\
PCVRPL      & $\surd$ & $\surd$ &   &   &   & $\surd$ \\
CVRPLS      & $\surd$ & $\surd$ &   &   & $\surd$ &   \\
PCVRPLS     & $\surd$ & $\surd$ &   &   & $\surd$ & $\surd$ \\
CVRPLTW     & $\surd$ & $\surd$ & $\surd$ &   &   &   \\
PCVRPLTW    & $\surd$ & $\surd$ & $\surd$ &   &   & $\surd$ \\
CVRPLSTW    & $\surd$ & $\surd$ & $\surd$ &   & $\surd$ &   \\
PCVRPLSTW   & $\surd$ & $\surd$ & $\surd$ &   & $\surd$ & $\surd$ \\
\hline
\end{tabular}%
}
\end{table}

%% file: Tabs/24VRP-Prompt.tex

\begin{table}[]
\centering
\caption{Examples of constraint descriptions for 24 VRP variants.}~\label{table:24_prompt}
\tiny
\renewcommand{\arraystretch}{1.35} 
\resizebox{0.99\textwidth}{!}{%
\begin{tabular}{|c|p{8.5cm}|}
\hline
\multicolumn{1}{|c|}{\textbf{Problem}} & \multicolumn{1}{c|}{\textbf{Problem Example}} \\
\hline
CVRP         & The total load of each route must not exceed the vehicle capacity.                                                                                                                                                                                                                                                                                                                                                                                                                                   \\ \hline
HCVRP        & The total load of each route must not exceed the vehicle capacity. Specifically, there should be at least 3 routes where the total load is less than 100 units.                                                                                                                                                                                                                                                                                                                                      \\ \hline
MCVRP        & The total load of each route must not exceed the vehicle capacity. Additionally, nodes [12, 14] require deliveries of [70, 80] units of a new type of goods. The maximum load capacity for this type of goods on each route is 100 units, and problem data excludes information about new goods.                                                                                                                                                                                                  \\ \hline
DCVRP        & The total load of each route must not exceed the vehicle capacity. Specifically, for node [19], its base demand is augmented by 5 times the square root of the accumulated travel distance from the depot [0] to that node.                                                                                                                                                                                                         \\ \hline
VRPL         & Each route must not exceed 150 units in length.                                                                                                                                                                                                                                                                                                                                                                                                                                                      \\ \hline
HVRPL        & Each route must not exceed 200 units in length, and at least three routes must have a total length of less than 150 units.                                                                                                                                                                                                                                                                                                                                                                           \\ \hline
RVRP         & After visiting node [17], the vehicle's remaining allowable travel distance for that route is reset to 150 units. At each node, the remaining driving distance cannot be negative.                                                                                                                                                                                                                                                                                                                   \\ \hline
DVRPL        & Each route must not exceed 200 units in length. The vehicle's remaining range decreases with each visit. After visiting node [17], the remaining range will be halved.                                                                                                                                                                                                                                                                                                                                                                               \\ \hline
VRPTW        & The arrival time at each node must meet its specified time window.                                                                                                                                                                                                                                                                                                                                                                                                                                   \\ \hline
HVRPTW       & The arrival time at each node must meet its specified time window. Specifically, one route must have its start time is 300, while all other routes start with time 0.                                                                                                                                                                                                                                                \\ \hline
VRPMTW       & The arrival time at each node must meet its specified time window. For node [4], in addition to its original time window, an additional time window of [900, 950] is also available.                                                                                                                                                                                                                                                                                                               \\ \hline
DVRPTW       & The arrival time at each node must meet its specified time window. For node [18], the service time dynamically increases by the amount of time from the start of its time window to the arrival time.                                                                                                                                                                                                                                                                                                \\ \hline
VRPMPD       & At Node [24], 10 units of goods are picked up. Unlike delivery, visiting a pickup node reduces the vehicle's available capacity.                                                                                                                                                                                                                                                                                                                                                                     \\ \hline
HVRPMPD      & At Node [24], 10 units of goods are picked up. Unlike delivery, visiting a pickup node reduces the vehicle's available capacity. Specifically, there should be at least 3 routes where the total load is less than 100 units.                                                                                                                                                                                                                                                                        \\ \hline
MVRPMPD      & Nodes [12, 14] require deliveries of [70, 80] units of a new type of goods. At Node [24], I pick up 20 units of these goods and 10 units of original goods. Before pick up, it needs to check whether sufficient goods have been delivered. Both types of goods are stored separately, with a maximum load of 100 units for new goods on each route, and problem data excludes information about new goods.  \\ \hline

DVRPMPD      & At Node [24], 10 units of goods are picked up, along with an additional amount calculated as 5 times the square root of the accumulated travel distance from the depot [0] to this node. Unlike delivery, visiting a pickup node reduces the vehicle's available capacity.                                                                                                                                  \\ \hline
VRPSVC       & Nodes [13, 23] must be on the same route.                                                                                                                                                                                                                                                                                                                                                                                                                                                            \\ \hline
CluVRP       & Nodes [7, 10] must be on the same route, and these nodes must be visited consecutively.                                                                                                                                                                                                                                                                                                                                                                 \\ \hline
VRPSO        & Nodes [13, 23] must be on the same route, and node [13] must be visited before node [23].                                                                                                                                                                                                                                                                                                                                                                                                            \\ \hline
VRPILC       & Nodes [7, 8] must not be on the same route.                                                                                                                                                                                                                                                                                                                                                                                                                                                          \\ \hline
PVRP         & Nodes [5, 7] are priority points.                                                                                                                                                                                                                                                                                                                                                                                                             \\ \hline
VRPRP        & Node [8] is the priority node and must be one of the first three positions in at least one route.                                                                                                                                                                                                                                                                                                                                                                                                   \\ \hline
VRPMP        & Nodes [7, 5, 3] are priority nodes with strictly decreasing priority levels: [7] (highest), [5], and [3] (lowest). Higher-priority nodes must be visited before lower-priority ones and other nodes.                                                                                                                                                                                                                                                                                             \\ \hline
VRP-dRP      & Nodes [7, 5, 3] follow the d-relaxed priority rule with decreasing priority: [7] (highest), [5], and [3] (lowest). Each node can be visited within its level or one level later, but no lower-priority node can be visited more than one level early. Other nodes are non-priority.                                                                                                                                                                                                             
\\ \hline
\end{tabular}%
}
\end{table}

%% file: 12-A-Base-models.tex
\newpage
\section{Function Template}\label{12-A-vio_temp}

The Constraint Scoring Program is designed to guide the generation of scoring functions in a structured manner. The following code template serves as a framework to define the scoring function, ensuring clarity and proper parameter usage. Within the template, the function name is predefined, and the roles of relevant parameters are described in detail. The specific implementation details are generated by the LLM based on the target VRP variant.

\customcode[1]{Base-model/vio_temp.tex} 

\section{Database}\label{12-A-Database}

The database contains six representative constraints, which exemplify problems related to these constraints. If no example problem is associated with the given input $I$, the base case $(I_0, C_0)$, which includes no relevant constraints, will be selected. The following sections introduce these six representative constraints and the base case.

\subsection{No Relevant Constraints}

\textbf{Constraint Description:} No relevant constraints.

\textbf{Verification Code:}

\customcode[1]{Base-model/Base.tex} 

\customcode[14]{Base-model/Base1.tex}

\subsection{Vehicle Capacity}~\label{12-A-Exp-C}

\textbf{Constraint Description:} The total load of each route must not exceed the vehicle capacity.

\textbf{Verification Code:}

\customcode[1]{Base-model/CVRP.tex}

\newpage

\subsection{Distance Limit}

\textbf{Constraint Description:} Each route's total length must not exceed 100 units.

\textbf{Verification Code:}

\customcode[1]{Base-model/VRPL.tex}

\newpage
\subsection{Time Windows}

\textbf{Constraint Description:} The arrival time at the node must meet the time windows.

\textbf{Verification Code:}

\customcode[1]{Base-model/VRPTW.tex}

\newpage
\subsection{Pickup and Delivery}

\textbf{Constraint Description:} At Node [24], 10 units of goods are picked up. Unlike delivery, visiting a pickup node reduces the vehicle's available capacity.

\textbf{Verification Code:}

\customcode[1]{Base-model/VRPMPD.tex} 

\newpage
\subsection{Same Vehicle}~\label{12-A-Exp-S}

\textbf{Constraint Description:} Nodes [1, 10] must be served by the same vehicle.

\textbf{Verification Code:}

\customcode[1]{Base-model/VRPS.tex}

\newpage
\subsection{Priority}

\textbf{Constraint Description:} Nodes [17, 19] are priority points.

\textbf{Verification Code:}

\customcode[1]{Base-model/PVRP.tex}

%% file: 13-A-Baselines_License.tex
\section{Licenses}\label{13-Appendix-Baselines+Licenses}

The licenses and URLs of the baseline methods are provided in Table~\ref{1}.

\begin{table}[H]
\centering
\setlength{\tabcolsep}{1mm}
\renewcommand\arraystretch{1.2}
\caption{A summary of licenses.}
\resizebox{\textwidth}{!}{
\begin{tabular}{llll}
\toprule
Resources         & Type    & URL                                                         & License                             \\ \midrule
CPLEX             & Code    & \url{https://www.ibm.com/products/ilog-cplex-optimization-studio} & Available for academic research use \\
OR-Tools          & Code    & \url{https://github.com/google/or-tools}                          & Apache License 2.0                  \\
Gurobi            & Code    & \url{https://www.gurobi.com/}                                     & Available for academic research use \\ \midrule
LKH-3             & Code    & \url{http://webhotel4.ruc.dk/~keld/research/LKH-3/}               & Available for academic research use \\
PyVRP-HGS         & Code    & \url{https://github.com/PyVRP/PyVRP}                              & MIT License                         \\ \midrule
Reflexion        & Code    & \url{https://github.com/noahshinn/reflexion}                      & MIT License                         \\
PHP               & Code    & \url{https://github.com/chuanyang-Zheng/Progressive-Hint}         & Available online                    \\
CoE               & Code    & \url{https://github.com/xzymustbexzy/Chain-of-Experts/tree/main}  & Available online                    \\
Self-verification & Code    & \url{https://github.com/Zhehui-Huang/LLM\_Routing}                & MIT License                         \\ \midrule
Solomon           & Dataset & \url{http://vrp.atd-lab.inf.puc-rio.br/index.php/en/}             & Available for academic research use \\
CVRPLib           & Dataset & \url{http://vrp.atd-lab.inf.puc-rio.br/}              & Available for academic research use\\ \bottomrule
\end{tabular}}
\label{1}
\end{table}